\pdfoutput=1
\documentclass{article}

\PassOptionsToPackage{numbers, compress}{natbib}

\usepackage[preprint]{neurips_2026}

\usepackage[utf8]{inputenc} 
\usepackage[T1]{fontenc}    
\usepackage{url}            
\usepackage{booktabs}       
\usepackage{amsfonts}       
\usepackage{nicefrac}       
\usepackage{microtype}      
\usepackage{xcolor}         

\usepackage{graphicx}
\usepackage{wrapfig}
\usepackage{multirow}
\usepackage[table]{xcolor}
\usepackage{enumitem}
\usepackage[tableposition=top]{caption}
\usepackage{float}
\usepackage{placeins}
\usepackage{amsmath}
\usepackage{algorithm}
\usepackage{algpseudocode}

\usepackage{hyperref}       
\usepackage[capitalize,nameinlink,noabbrev]{cleveref} 

\crefname{section}{Sec.}{Secs.}
\Crefname{section}{Section}{Sections}

\crefname{figure}{Fig.}{Figs.}
\Crefname{figure}{Figure}{Figures}

\crefname{table}{Tab.}{Tabs.}
\Crefname{table}{Table}{Tables}

\crefname{equation}{Eq.}{Eqs.}
\Crefname{equation}{Equation}{Equations}

\usepackage{xspace}

\makeatletter
\DeclareRobustCommand\onedot{\futurelet\@let@token\@onedot}
\def\@onedot{\ifx\@let@token.\else.\null\fi\xspace}
\def\eg{\textit{e.g}\onedot} 
\def\ie{\textit{i.e}\onedot}

\makeatother

\title{FuTCR: Future-Targeted Contrast and Repulsion for Continual Panoptic Segmentation}

\author{
Nicholas Ikechukwu\thanks{Equal contribution} \quad
Keanu Nichols\footnotemark[1] \quad
Deepti Ghadiyaram \quad
Bryan A. Plummer \\
Boston University \\
\texttt{\{ncholas, kmn5409, dghadiya, bplum\}@bu.edu}
}

\begin{document}

\maketitle

\begin{abstract}
Continual Panoptic Segmentation (CPS) requires methods that can quickly adapt to new categories over time. The nature of this dense prediction task means that training images may contain a mix of labeled and unlabeled objects. As nothing is known about these unlabeled objects \textit{a priori}, existing methods often simply group any unlabeled pixel into a single "background" class during training.  In effect, during training, they repeatedly tell the model that all the different background categories are the same (even when they aren't).  This makes learning to identify different background categories as they are added challenging since these new categories may require using information the model was previously told was unimportant and ignored. Thus, we propose a Future-Targeted Contrastive and Repulsive (FuTCR) framework that addresses this limitation by restructuring representations before new classes are introduced. FuTCR first discovers confident \textit{future-like} regions by grouping model-predicted masks whose pixels are consistently classified as background but exhibit non-background logits. Next, FuTCR applies pixel-to-region contrast to build coherent prototypes from these unlabeled regions, while simultaneously repelling background features away from known-class prototypes to explicitly reserve representational space for future categories. Experiments across six CPS settings and a range of dataset sizes show FuTCR improves relative new-class panoptic quality over the state-of-the-art by up to \textbf{28\%}, while preserving or improving base-class performance with gains up to \textbf{4\%}\footnote{The models and code will be made publicly available:\emph{ \url{https://github.com/futcr/FuTCR}}}. 

\end{abstract}

\section{Introduction}
\label{sec:intro}

\begin{figure*}[t]
    \centering
    \includegraphics[width=0.85\textwidth]{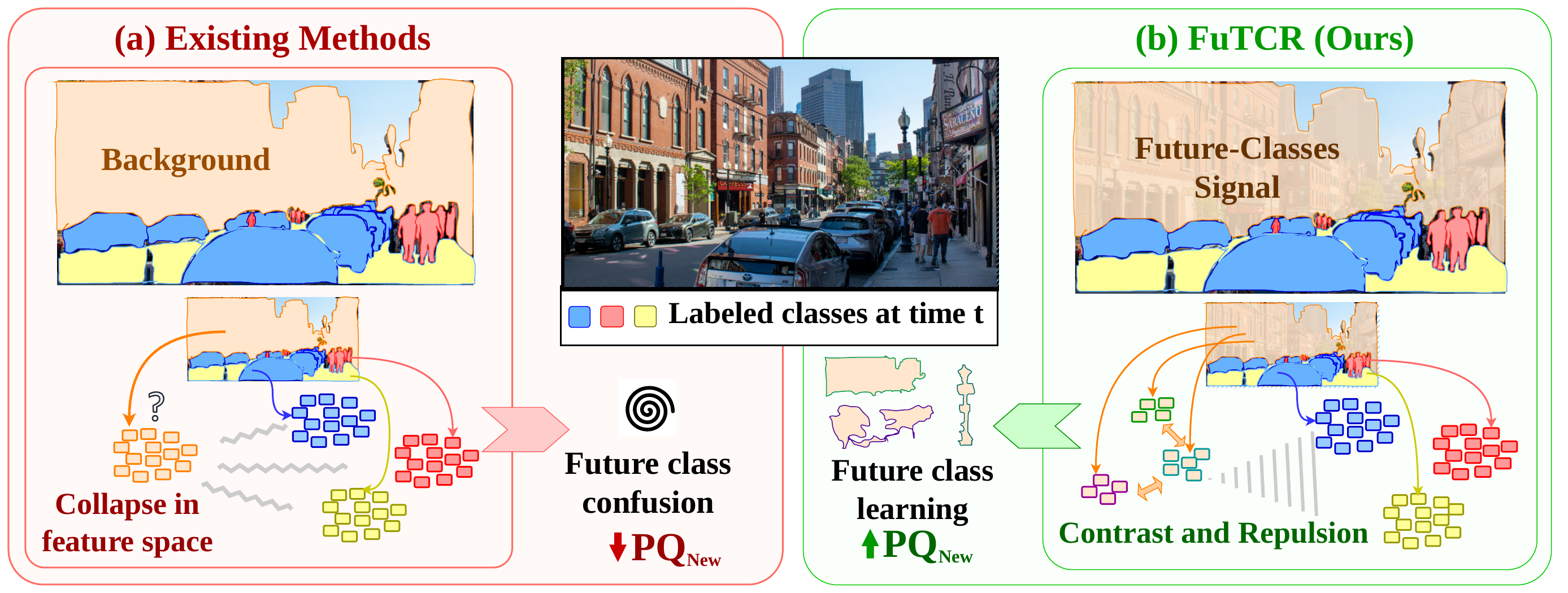}
    \caption{
    \textbf{From background noise to structured supervision}. (a) Prior continual panoptic methods treat background regions as non-informative~\cite{fang2025combo,zhu2025rethinking,cermelli2023comformer,chen2024strike}, allowing future-class evidence to be absorbed into existing decision regions in feature space over time. (b) Our framework instead converts background activations into future-aware structural cues that organize scene composition before new labels arrive, reducing semantic drift and improving Panoptic Quality (PQ).}
     \label{fig:motivation_ours}
      \vspace{-8pt}
\end{figure*}

Continual learning aims to expand a model’s recognition capability by introducing new categories over time without retraining from scratch~\cite{zhou2025ferret,kang2025your,castro2018end,rebuffi2017icarl,chaudhry2018riemannian,shin2017continual,yan2021dynamically,tao2020few,jiao2024vector,wang2021ordisco}. Most prior work studies classification, where the model updates only a global label predictor~\cite{kang2025your,jiao2024vector,friedman2026pacbayes,ye2025dynamic,wu2025exploiting,momeni2025anacp,chenyanxi2026hippotune}, where samples only contain examples of labeled classes. In contrast, dense prediction tasks like image segmentation may contain a mix of labeled and unlabeled categories~\cite{yang2025adapt,yin2025beyond,truong2025falcon,zhu2025continual,zhang2025parameter,fang2025combo, zhu2023continual,zhu2025rethinking, tang2023contrastive, shi2024devil, zhang2023coinseg,Lin_2025_CVPR_universal_segmen,kim2024eclipse}. As shown in \cref{fig:motivation_ours}(a), many of these methods map all unlabeled pixels into a single "background" category. Thus, these methods may only learn how to identify it as \textit{any} unlabeled category, rather than learning how to distinguish between objects.  For example, if no vehicles are labeled, then a model could just look for a wheel to mark that object as background. Thus, there would be no need to learn to separate cars and trucks even though they may appear as new categories to learn in the future.

We address this limitation by proposing a \textbf{Fu}ture-\textbf{T}argeted \textbf{C}ontrastive and \textbf{R}epulsive (\textbf{FuTCR}) learning framework that explicitly restructures the representation space \textit{before} new classes are introduced. As illustrated in \cref{fig:motivation_ours}(b), our core idea is to treat unlabeled content that is likely to belong to future classes not as undifferentiated background, but as structured regions that should remain separable from known-class representations. To do so, FuTCR uses the model’s own panoptic predictions at the base step to discover \textit{future-like} regions, \ie, connected mask regions that are consistently predicted as background yet exhibit non-background evidence in the logits. Within each discovered region, FuTCR applies a \textbf{pixel-to-region contrastive objective} that pulls pixels toward a shared region prototype and pushes them away from prototypes of other regions. This objective enforces coherence inside each unlabeled region and diversity across regions, encouraging the model to form distinct, reusable prototypes for future categories instead of collapsing them into background.

To further prevent unlabeled regions from being absorbed into known-class decision regions, FuTCR introduces a \textbf{future-targeted repulsion objective}. Concretely, we maintain a set of prototypes for known classes and explicitly repel features from discovered future-like regions away from these prototypes in both feature and logit space. This repulsion discourages future evidence from occupying the same representational subspace as old classes and creates a margin around known-class prototypes that can later be allocated to new categories. Together, the contrastive and repulsive terms cause the current-step model to pre-organize its feature space so that future-class pixels form coherent clusters that are distinct from old classes, which eases the integration of new categories during incremental updates. Our future-aware error and representation analysis
confirms that FuTCR significantly reduces future-old class confusion compared to strong baselines.

While prior work has explored using unlabeled data via semi-supervised or self-supervised objectives~\cite{sohn2020fixmatch,chen2020simple,he2020momentum,olsson2021classmix}, these methods primarily improve performance on currently labeled classes and do not explicitly shape the feature space for \textbf{future class expansion} in continual segmentation. Recent work on future-aware representation learning for segmentation~\cite{lin2023preparing} moves in this direction but relies on externally defined regions or non-adaptive clustering, which decouples region construction from the model’s evolving predictions. In contrast, FuTCR couples region discovery, representation learning, and classifier preparation in a single model-driven procedure: predicted masks define the regions, contrastive learning structures the corresponding feature clusters, and repulsion from known-class prototypes reserves capacity for future categories. This tight coupling yields a scalable, adaptive mechanism for organizing representations ahead of future class introductions,  resulting in a relative gain of up to \textbf{28\%} on new classes coupled with gains up to \textbf{4\%} on old categories.

In summary, our main contributions are:
\begin{itemize}[nosep,leftmargin=*]
    \item We identify a fundamental \textbf{extensibility limitation} in continual panoptic segmentation: representations and classifier parameters become progressively biased toward previously learned categories, suppressing future-class integration under background overlap.
    
     \item We propose \textbf{FuTCR}, a \textbf{future-targeted region-level unsupervised learning framework} that leverages model-predicted masks to construct supervision from unlabeled regions, combining \textbf{contrastive structuring} and \textbf{known-class repulsion} to reserve explicit capacity for future categories.  
      \item Through a dedicated future-aware error and representation analysis, we show that FuTCR reshapes the feature space to reduce future-known class confusion, leading to consistent gains in novel-class panoptic quality across two extreme settings, including fully shared and fully disjoint image overlaps, while preserving performance on previously learned classes.
    
\end{itemize}

\section{Related Work}
\label{sec:related_work}

Recent continual panoptic segmentation methods typically adapt dense prediction architectures to class-incremental streams by combining mask-based distillation~\cite{fang2025combo, zhu2025rethinking, yuan2024continual}, pseudo-labeling~\cite{cermelli2023comformer,yuan2024continual}, replay~\cite{chen2024strike, cermelli2023comformer, cha2021ssul}, or prompt tuning~\cite{kim2024eclipse, manjunath2025vista} to alleviate catastrophic forgetting~\cite{french1999catastrophic,robins1995catastrophic,thrun1998lifelong} and maintain panoptic quality ~\cite{cermelli2023comformer,fang2025combo,chen2024strike,zhu2025rethinking,kim2024eclipse,yuan2024survey}. These approaches successfully preserve performance on previously learned classes, but they still train the base step under a closed-world assumption: pixels from unseen categories are mapped to background and classifier capacity is allocated only to currently labeled classes~\cite{fang2025combo,zhu2025rethinking,park2024mitigating,cermelli2020modelingMIB, yuan2024continual}. As a result, future-class evidence is repeatedly absorbed into background representations, biasing decision regions toward old categories and suppressing novel-class predictions when new labels arrive~\cite{yuan2024survey,kalb2024principles,kim2024eclipse}. FuTCR directly targets this structural extensibility limitation by treating background pixels as carriers of future-class signal and reshaping the feature space around them via future-targeted contrast and repulsion, rather than relying solely on supervised losses for current classes.

A complementary line of work argues that continual segmentation should be future-aware and distribution-aware, exploiting unlabeled pixels as hints about upcoming classes instead of discarding them~\cite{lin2023preparing,lin2022continual,song2023scale,elaraby2024bacs}. Prior methods pre-learn future knowledge by applying contrastive or distillation objectives on unlabeled regions and show that completing or suppressing unlabeled foreground significantly affects forgetting~\cite{yin2025beyond,lin2023preparing,luo2025trace,park2024mitigating}.  However, these methods focus on semantic segmentation, that simply labels pixels as their category, whereas we focus on the panoptic segmentation task that gives \textit{pixel and instance} labels.  Further, unlike our approach none of these methods explicitly reserve classifier capacity for class expansion.  Open-set and open-world panoptic segmentation further treat unlabeled regions as unknown classes and maintain explicit unknown categories or discovery mechanisms~\cite{cha2021ssul,hwang2021exemplar,yin2024revisiting, dong2022region, xu2022dual}, yet these formulations are typically single-step and do not exploit background pixels to proactively structure feature space for future categories across a continual stream. FuTCR closes this gap by coupling model-predicted masks with future-targeted contrast and repulsion on background pixels inside a query-based panoptic framework, carving out future-ready, non-overlapping representational regions that significantly enhance novel-class panoptic quality in both fully overlapped and fully disjoint incremental settings.
  
\section{Future-Targeted Contrastive and Repulsive (FuTCR) Framework}
\label{sec:method}

We explore an incremental panoptic segmentation setting defined over a fixed semantic label space $\mathcal{C} = \{1,\dots,K\}$, where $K$ is the total number of semantic classes to be learned across all steps (seen and unseen). Training proceeds over $T$ steps. At a given time step $t \in \{1,\dots,T\}$, the learner receives a dataset $\mathcal{D}_t = \{(x, y)\}$, where $x$ is an input image, $y$ is its panoptic annotation containing both \textit{thing} instances and \textit{stuff} regions for the current class set $\mathcal{C}^t \subseteq \mathcal{C}$, and $\mathcal{Y}(x) \subseteq \mathcal{C}$ denotes the subset of classes present in image $x$. The cumulative set of classes seen up to step $t$ is $\mathcal{C}^{\leq t} = \bigcup_{s=1}^{t} \mathcal{C}^s$. The goal of incremental panoptic segmentation is: after training on $\mathcal{D}_t$, the model $f_{\theta_t}$ must accurately segment all classes in $\mathcal{C}^{\leq t}$, despite only having direct access to $\mathcal{D}_t$ and not previous steps.

Following~\cite{zhu2025rethinking}, we use a query-based panoptic architecture. Let $Q$ denote the number of queries and $d$ the feature dimension. For an image $x$ at step $t$, the decoder produces per-query feature representations $\mathbf{h}_t(x) \in \mathbb{R}^{Q \times d}$. A shared classifier (prediction head) is parameterized by weights $W_t \in \mathbb{R}^{K \times d}$, mapping query features to class scores over the entire label space $\mathcal{C}$, including categories that are not yet annotated at step $t$ but may already occur in $\mathcal{Y}(x)$.

In the default training protocol, supervision at step $t$ is restricted to the current classes $\mathcal{C}^t$, while pixels from unseen classes $\mathcal{C} \setminus \mathcal{C}^{\leq t}$ are assigned to a generic background label. Consequently, the query features $\mathbf{h}_t(x)$ and classifier $W_t$ are optimized only to separate $\mathcal{C}^{\leq t}$ from background. Future-class evidence is thus repeatedly collapsed into the background subspace and shaped to sit near already-seen classes in order to separate them from non-object regions, discarding structure that could benefit later steps, as illustrated in~\cref{fig:motivation_ours}. We tackle this limitation in FuTCR as detailed next.

\begin{figure*}[t]
    \centering
    \includegraphics[width=.9\textwidth]{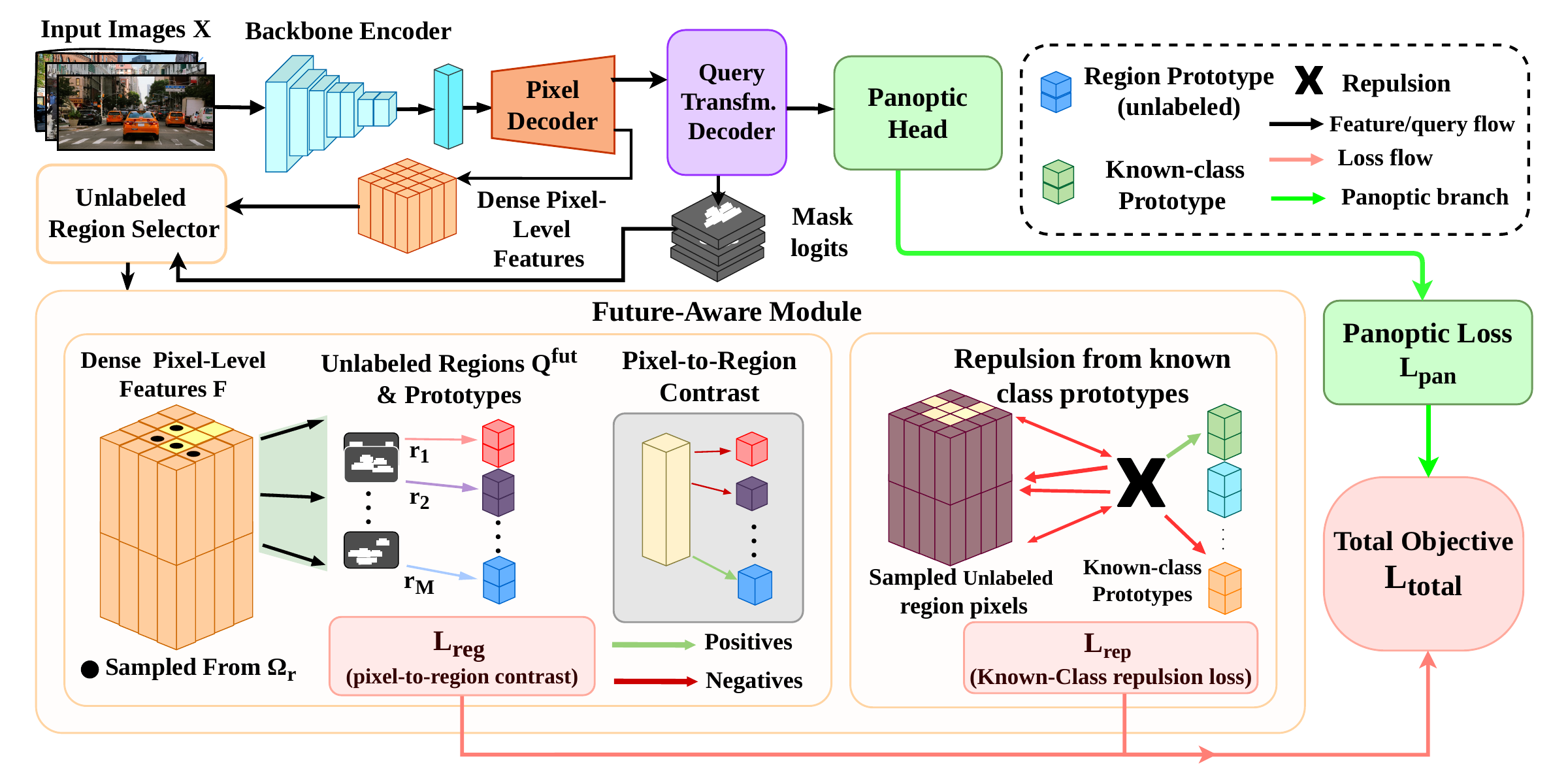}
   \caption{
    Overview of \textbf{FuTCR}, a query-based continual panoptic model produces dense region features and mask predictions. 
    Our future-targeted module leverages unlabeled regions together with ground-truth labeled regions to perform \textbf{unlabeled region discovery}~\cref{subsec:future_regions}, \textbf{region-level contrastive learning}~\cref{subsec:region_contrast}, and \textbf{known-class repulsion}~\cref{subsec:known_class_repulsion}, encouraging structured representations for future categories. 
    The final objective combines standard panoptic losses with future-targeted contrastive and repulsion objectives~\cref{eq:total_loss}. 
    }
    \label{fig:method}
    \vspace{-8pt}

\end{figure*}

\subsection{Unlabeled Region Discovery}
\label{subsec:future_regions}

During training pixels from classes in $\mathcal{C} \setminus \mathcal{C}^{\leq t}$ already appear in $x$ at step $t$, but prior work maps them to a generic background label and they never receive class-specific supervision (\eg,~\cite{cermelli2020modelingMIB,fang2025combo,zhu2025rethinking,lin2023preparing}). Our aim is to convert this systematically ignored evidence into an explicit structural signal at the base step to produce a representation space that already allocates coherent regions for future classes before they are introduced.

For a given image $x_b$ and query index $q$ panoptic model $f_{\theta_t}$ produces per-query mask $\hat{m}_{b,q} \in [0,1]^{H \times W}$. We then identify \emph{future-like} regions whose support lies predominantly on pixels outside the current class set $\mathcal{C}^{\leq t}$ (e.g., unlabeled labels in the ground truth or pixels annotated as background but not belonging to $\mathcal{Y}(x_b) \cap \mathcal{C}^{\leq t}$). For each image index $b$, we define the set of future queries
\[
\mathcal{Q}_b^{\mathrm{fut}}
=
\left\{
q \;\middle|\;
\begin{array}{l}
\hat{m}_{b,q} \text{ is confident and sufficiently large, and}\\\\[-1pt]
\text{a majority of its support lies on pixels outside } \mathcal{C}^{\leq t}
\end{array}
\right\},
\]
and collect batch-level future regions as $\mathcal{R}^{\mathrm{fut}} = \{(b,q) \mid q \in \mathcal{Q}_b^{\mathrm{fut}}\}$. This isolates step $t$'s object-like masks whose identity is unknown but whose pixels may later be assigned to classes in $\mathcal{C} \setminus \mathcal{C}^{\leq t}$.

Unlike prior methods that collapse future evidence into a single auxiliary "unknown" class~\cite{cermelli2020modelingMIB} or depend on externally generated region proposals~\cite{lin2023preparing,kirillov2019panoptic}, where discovered regions are only indirectly aligned with the incremental classifier $W_t$, our approach is fully model-driven. FuTCR derives $\mathcal{R}^{\mathrm{fut}}$ directly from the query masks of $f_{\theta_t}$, preserving the panoptic head’s native one-query-per-region structure. This tight coupling grounds future region discovery in the same query feature space reused by subsequent heads, yielding a more faithful and scalable foundation for future-targeted contrastive and repulsive learning~\cite{zheng2021contrastive, bohm2022attraction, yu2001segmentation}.

\subsection{Region Prototypes and Pixel-to-Region Contrast}
\label{subsec:region_contrast}

For each future-like region $\in \mathcal{R}^{\mathrm{fut}}$, we pool corresponding mask features to form region prototypes. Let $F_b \in \mathbb{R}^{C \times H \times W}$ denote image $x_b$'s mask feature map, then prototype for region $r = (b,q)$ is
\begin{equation}
\mathbf{p}_r
=
\frac{1}{|\Omega_r|}
\sum_{(i,j) \in \Omega_r}
F_b(:, i, j),
\quad
\Omega_r
=
\{(i,j) \mid \hat{m}_{b,q}(i,j) > \tau_{\mathrm{mask}}\},
\label{eq:region_proto}
\end{equation}
where $\tau_{\mathrm{mask}}$ is a confidence threshold. This simple averaging ties each prototype directly to a single panoptic query, so that future categories are represented as object-level clusters in the same feature space that the incremental classifier will later reuse. We then sample a fixed number of pixel features from each $\Omega_r$ to serve as anchors $\mathbf{f}_n \in \mathbb{R}^C$, each associated with its region index $r(n)$.

To encourage future-like regions to form coherent object-centric representations while remaining distinct from one another, we apply an InfoNCE-style contrastive loss~\cite{chen2020simple, he2020momentum} between sampled pixel features and region prototypes. Let $\mathcal{P}$ denote the set of all prototypes in the batch. For each sampled pixel feature $\mathbf{f}_n$ with positive prototype $\mathbf{p}_{r(n)}$, we define
\begin{equation}
\mathcal{L}_{\mathrm{reg}}
=
-\frac{1}{N}
\sum_{n=1}^{N}
\log
\frac{
\exp\big(\mathrm{sim}(\mathbf{f}_n, \mathbf{p}_{r(n)}) / \tau\big)
}{
\sum_{\mathbf{p}_k \in \mathcal{P}}
\exp\big(\mathrm{sim}(\mathbf{f}_n, \mathbf{p}_k) / \tau\big)
},
\label{eq:region_infonce}
\end{equation}
where $\mathrm{sim}(\cdot,\cdot)$ is cosine similarity and $\tau$ is a temperature. This objective encourages pixels from the same future-like region to cluster around a shared prototype.  Further, prototypes for different regions are pushed apart so future-class evidence is organized into stable, query-aligned object capsules instead of being spread diffusely across background, which we will show is a failing of prior work.

\subsection{Known-class Repulsion}
\label{subsec:known_class_repulsion}

Future-like regions are composed largely of unlabeled pixels in the current step’s ground truth, which standard continual methods either drop from the loss or treat as background. To prevent these pixels from collapsing onto known-class decision regions, we introduce an known-class-repulsion loss that explicitly pushes unlabeled features away from prototypes of known classes $\mathcal{C}^{\leq t}$, thereby reserving representational “headroom” for future categories.

Let $\mathcal{I}^{\mathrm{unlb}}$ be a set of sampled unlabeled pixels across the batch and let $\{\mathbf{w}_c\}_{c \in \mathcal{C}^{\leq t}}$ denote the normalized classifier prototypes or class-centric feature vectors for known classes at step $t$. For each unlabeled feature $\mathbf{z}_u$ and each known class $c$, we compute the cosine similarity $s_{u,c} = \mathrm{sim}(\mathbf{z}_u,\mathbf{w}_c)$ and apply a margin-based hinge penalty:
\begin{equation}
\mathcal{L}_{\mathrm{rep}}
=
\frac{1}{|\mathcal{I}^{\mathrm{unlb}}|}
\sum_{u \in \mathcal{I}^{\mathrm{unlb}}}
\max\big(0,\; s_{u,c^{\star}(u)} - \gamma \big),
\label{eq:known_class_repulsion}
\end{equation}
where  $u \in I^{unlb}$, $c^{\star}(u) = \arg\max_{c \in \mathcal{C}^{\leq t}} s_{u,c}$ is the most similar known class and $\gamma$ is a repulsion margin. Intuitively, unlabeled pixels that align too strongly with any known class incur a positive loss and are pushed to move away in feature space, reducing the future-old class confusion we quantify in~\cref{subsec:future_aware_analysis}. This makes it less likely that new classes will later be forced into already-occupied decision regions when they are labeled.

\noindent\textbf{Overall Objective and Training Scheme.}

Our future-aware module is attached to a query-based panoptic backbone (\eg, a SimCIS-style~\cite{zhu2025rethinking} transformer decoder) and optimized jointly with the standard panoptic objective. Let $\mathcal{L}_{\mathrm{pan}}$ denote the conventional segmentation, classification, and localization losses for the current classes $\mathcal{C}^t$ (see~\cite{zhu2025rethinking}). The overall objective at step $t$ is
\begin{equation}
\mathcal{L}_{\mathrm{total}}
=
\mathcal{L}_{\mathrm{pan}}
+
\lambda_{\mathrm{reg}} \,\mathcal{L}_{\mathrm{reg}}
+
\lambda_{\mathrm{rep}} \,\mathcal{L}_{\mathrm{rep}},
\label{eq:total_loss}
\end{equation}
where $\lambda_{\mathrm{reg}}$ and $\lambda_{\mathrm{rep}}$ weight future-targeted region contrast and known-class repulsion, respectively. Both objectives operate on pixels that standard continual methods typically ignore or absorb into background~\cite{cermelli2020modelingMIB,park2024mitigating,yin2025beyond}, encouraging future-like regions to form coherent representations separated from known-class decision regions. In the next section we will show pre-structured feature space improves future-class integration under challenging overlap regimes while preserving stable panoptic predictions for previously learned categories.

\section{Experiments}
\label{sec:experiments}

Following~\cite{yang2025adapt,fang2025combo,zhu2025rethinking,cermelli2023comformer}, we evaluate FuTCR on diverse continual panoptic segmentation settings derived from ADE20K~\cite{zhou2017sceneade20k}. The full label space $\mathcal{C}=\{1,\ldots,K\}$ contains $K$ semantic classes and is partitioned across $T$ continual learning steps into a base class set $\mathcal{C}^1$ and a sequence of incremental class groups $\{\mathcal{C}^t\}_{t=2}^{T}$. At step $t=1$, the model is trained only on $\mathcal{D}_1$ with annotations for $\mathcal{C}^{1}$. At step $t>1$, a new group $\mathcal{C}^{t}$ is introduced and supervision is restricted to the current classes $\mathcal{C}^{t}$, while evaluation is always performed on all classes seen so far, $\mathcal{C}^{\le t} = \bigcup_{s=1}^{t} \mathcal{C}^{s}$. In practice, $\mathcal{D}_1$ has $\sim$10K images and incremental steps have $\sim$700 images on average. The validation and test sets has 1K held-out images each. \emph{New classes} are those introduced after the base step. Thus, at step $t$ the new classes observed so far are $\mathcal{C}^{2:t}$, and at the final step they are $\mathcal{C}^{2:T}$. Similarly, \emph{future classes} are defined from the perspective of step $t$ and refers to classes that have not yet been introduced, $\mathcal{C}^{t+1:T}$.

In all cases, the base step uses $|\mathcal{C}^{1}| = 100$ classes and the remaining $50$ classes in $\mathcal{C}^{2:T}$ are introduced over $T-1$ incremental steps in groups of $k \in \{5,10,50\}$ classes per step, resulting in $T = \{11, 6, 2\}$, respectively. 
We report final-step performance after incorporating all incremental classes. Specifically, PQ$_\text{base}$ is computed on the base classes $\mathcal{C}^1$, PQ$_\text{new}$ is computed on the incremental classes $\mathcal{C}^{2:T}$, and PQ$_\text{all}$ is computed on all classes $\mathcal{C}^{1:T}$.

 Each of the cases above are evaluated under two settings. The \emph{overlap stream} reuses images from step $1$ in later steps. As such, pixels belonging to all future classes in $\mathcal{C} \setminus \mathcal{C}^{\le t}$ are present in $x \in \mathcal{D}_t$, but annotated only as background or ignored in prior work as discussed earlier. This stream directly tests whether a method can prevent future-class evidence from collapsing into background or being confused with known classes when future classes are visually present but unlabeled. The \emph{disjoint-image stream} contains no image overlaps between the base and incremental steps. This simulates a setting where new classes are added to a pretrained model without access to the original training data.  We note that evaluating both streams makes our protocol more robust than some recent methods, which only use one (typically the overlap stream)~\cite{zhu2025rethinking,fang2025combo,chen2024strike,manjunath2025vista}.

\textbf{Metrics.} We adapt the same metrics as prior work~\cite{kim2024eclipse,zhu2025rethinking,fang2025combo}. Specifically, we report Panoptic Quality (PQ), which balances both recognition and segmentation quality to evaluate continual panoptic segmentation performance. In addition, we report the mean Intersection-over-Union (mIoU) computed between the ground truth and predicted segmentation maps. Eclipse reports PQ under our evaluation protocol, but its released code does not directly output mIoU; therefore, we report only PQ for Eclipse.

\textbf{Implementation Details}. Our method is implemented on top of Mask2Former~\cite{cheng2022masked} using the Detectron2 framework, with a pretrained ResNet-50 backbone~\cite{he2016deepresnet}. Unless otherwise specified, we follow the same training setup and hyperparameters as prior work~\cite{zhu2025rethinking,kim2024eclipse} to ensure fair comparison. We train the model sequentially across incremental steps, initializing each step from the previous model. We set the repulsion factor   $\lambda_{rep}$ for our future-aware module to $0.5$. All experiments were run on 4 NVIDIA RTX A6000 GPUs. Each incremental training step required approximately 2 hours on average for the $100$ - $5$ setting, and models were trained sequentially across all incremental steps. All reported baselines are evaluated under the same controlled split construction, class order, validation/test split, and final-step evaluation protocol. Additional implementation details are provided in \cref{app:split_construction} including compute details, training schedules, and hyperparameters.

\subsection{Results}

\cref{tab:main_results_100-5,tab:main_results_100-10,tab:main_results_100-50} reports both image stream's performance across three step sizes, resulting in six total combinations. On average FuTCR reports absolute gains of up to 4 points. Overall, our method is most advantageous when there are many steps (\cref{tab:main_results_100-5}), suggesting FuTCR becomes more advantageous over time (further supported by results in \cref{subsec:future_aware_analysis}). Notably, in PQ$_\text{new}$ \cref{tab:main_results_100-5} shows a $+28\%$ relative gain ($17.3\rightarrow 22.3$) over SimCIS~\cite{zhu2025rethinking}, the closest method to ours.  We also observe that our gains are not limited to just improvements in new classes.  For example, in the disjoint-image stream in \cref{tab:main_results_100-5} has most of its PQ gains on base classes, suggesting that the benefits can also stem from less destructive learning as we add base classes.  

\begin{table}[tb]
\caption{Performance on ADE20K \textbf{100-5} (11 total steps) under controlled overlap and disjoint-image streams. FuTCR improves performance by 2-3 points on average on PQ and mIOU metrics.}
\footnotesize
\setlength{\tabcolsep}{2pt}
\label{tab:main_results_100-5}
\centering
\begin{tabular}{lcccccccccccccc}
\toprule
 & \multicolumn{6}{c}{\textbf{Overlap}} & \multicolumn{6}{c}{\textbf{Disjoint}} & \multicolumn{2}{c}{\multirow{2}{*}{\textbf{Avg (all)}}} \\
& \multicolumn{2}{c}{1-100} & \multicolumn{2}{c}{101-150} & \multicolumn{2}{c}{all} & \multicolumn{2}{c}{1-100} & \multicolumn{2}{c}{101-150} & \multicolumn{2}{c}{all} & \\
\cmidrule(lr){2-3}\cmidrule(lr){4-5}\cmidrule(lr){6-7}\cmidrule(lr){8-9}\cmidrule(lr){10-11}\cmidrule(lr){12-13}\cmidrule(lr){14-15}
\textbf{Method} &  PQ$_\text{base}$ & mIOU & PQ$_\text{new}$ & mIOU & PQ$_\text{all}$ & mIOU &  PQ$_\text{base}$ & mIOU & PQ$_\text{new}$ & mIOU & PQ$_\text{all}$ & mIOU & PQ$_\text{avg}$ & mIOU\\
\midrule
Eclipse~\cite{kim2024eclipse}           & 37.5 & - & 8.5 & - & 28.8 & - & 36.6 & - & 7.1 & - & 27.0 & - & 27.9 & \\
Balconpas~\cite{chen2024strike}          & 35.0 & 41.3 & \textbf{22.3} & 21.6 & 30.8 & 34.8
& 34.2 & 40.6 & 20.6 & 20.9 & 29.6 & 34.1
& 30.2 & 34.4 \\
Combo~\cite{fang2025combo}           & 32.6 & 41.5 & 19.3 & 20.2 & 28.2 & 34.4
& 31.2 & 39.7 & 19.3 & 21.0 & 27.3 & 33.5
& 27.7 & 33.9 \\
SimCIS~\cite{zhu2025rethinking}           & 37.2 & 52.5 & 17.3 & 24.3 & 30.6 & 43.7
& 34.6 & 48.1 & 22.8 & 33.9 & 30.6 & 43.6
& 30.6 & 43.6 \\
\midrule

\textbf{FuTCR (Ours)}           & \textbf{38.8} & \textbf{53.7}
& \textbf{22.3} & \textbf{31.8}
& \textbf{33.3} & \textbf{46.8}
& \textbf{35.4} & \textbf{49.9}
& \textbf{23.0} & \textbf{37.3}
& \textbf{31.3} & \textbf{46.1}
& \textbf{32.3} & \textbf{46.4} \\

\bottomrule
\end{tabular}%
\end{table}

\begin{table}[tb]
\caption{Performance on ADE20K \textbf{100-10} (6 total steps) under controlled overlap and disjoint-image streams. FuTCR provides more stable results across both reported metrics, unlike prior work (\eg, Combo has similar average PQ but has an 8 point drop in mIOU).}
\footnotesize
\setlength{\tabcolsep}{2pt}
\label{tab:main_results_100-10}
\centering
\begin{tabular}{lcccccccccccccc}
\toprule
 & \multicolumn{6}{c}{\textbf{Overlap}} & \multicolumn{6}{c}{\textbf{Disjoint}} & \multicolumn{2}{c}{\multirow{2}{*}{\textbf{Avg (all)}}} \\
& \multicolumn{2}{c}{1-100} & \multicolumn{2}{c}{101-150} & \multicolumn{2}{c}{all} & \multicolumn{2}{c}{1-100} & \multicolumn{2}{c}{101-150} & \multicolumn{2}{c}{all} & \\
\cmidrule(lr){2-3}\cmidrule(lr){4-5}\cmidrule(lr){6-7}\cmidrule(lr){8-9}\cmidrule(lr){10-11}\cmidrule(lr){12-13}\cmidrule(lr){14-15}
\textbf{Method} &  PQ$_\text{base}$ & mIOU & PQ$_\text{new}$ & mIOU & PQ$_\text{all}$ & mIOU &  PQ$_\text{base}$ & mIOU & PQ$_\text{new}$ & mIOU & PQ$_\text{all}$ & mIOU & PQ$_\text{avg}$ & mIOU\\
\midrule

Eclipse~\cite{kim2024eclipse}           & 37.5 & -- & 10.8 & -- & 28.8 & --
& 37.7 & -- & 11.2 & -- & 29.1 & --
& 28.9 & -- \\
Balconpas~\cite{chen2024strike}         & 37.9 & 45.5 & 22.5 & 27.2 & 32.8 & 39.5
& 38.1 & 45.0 & 23.9 & 27.2 & \textbf{33.4} & 39.1
& 33.1 & 39.3 \\
Combo~\cite{fang2025combo}           & 37.4 & 44.9 & \textbf{25.6} & 28.9 & 33.4 & 39.6
& 36.6 & 44.0 & \textbf{24.4} & 28.7 & 32.5 & 38.9
& 32.9 & 39.2 \\
SimCIS~\cite{zhu2025rethinking}           & 39.9 & \textbf{55.7} & 20.1 & 26.8 & 33.3 & 46.5
& 37.8 & \textbf{54.9} & 19.1 & 31.2 & 31.6 & 47.4
& 32.4 & 46.9 \\
\midrule

\textbf{FuTCR (Ours)}           & \textbf{40.0} & 55.3
& 21.0 & \textbf{30.0}
& \textbf{33.7} & \textbf{47.0}
& \textbf{39.1} & 54.3
& 20.3 & \textbf{32.7}
& 32.8 & \textbf{47.5}
& \textbf{33.2} & \textbf{47.2} \\

\bottomrule
\end{tabular}%
\end{table}

\begin{table}[tb]
\caption{Performance on ADE20K \textbf{100-50} (2 total steps) under controlled overlap and disjoint-image streams. FuTCR improves performance by nearly 1 point on average on PQ and mIOU metrics.}
\footnotesize
\setlength{\tabcolsep}{2pt}
\label{tab:main_results_100-50}
\centering
\begin{tabular}{lcccccccccccccc}
\toprule
 & \multicolumn{6}{c}{\textbf{Overlap}} & \multicolumn{6}{c}{\textbf{Disjoint}} & \multicolumn{2}{c}{\multirow{2}{*}{\textbf{Avg (all)}}} \\
& \multicolumn{2}{c}{1-100} & \multicolumn{2}{c}{101-150} & \multicolumn{2}{c}{all} & \multicolumn{2}{c}{1-100} & \multicolumn{2}{c}{101-150} & \multicolumn{2}{c}{all} & \\
\cmidrule(lr){2-3}\cmidrule(lr){4-5}\cmidrule(lr){6-7}\cmidrule(lr){8-9}\cmidrule(lr){10-11}\cmidrule(lr){12-13}\cmidrule(lr){14-15}
\textbf{Method} &  PQ$_\text{base}$ & mIOU & PQ$_\text{new}$ & mIOU & PQ$_\text{all}$ & mIOU &  PQ$_\text{base}$ & mIOU & PQ$_\text{new}$ & mIOU & PQ$_\text{all}$ & mIOU & PQ$_\text{avg}$ & mIOU\\
\midrule

Eclipse~\cite{kim2024eclipse}           & 37.9 & -- & 15.9 & -- & 30.8 & --
& 38.1 & -- & 17.4 & -- & 31.2 & --
& 31.0 & -- \\
Balconpas~\cite{chen2024strike}         & 39.9 & 46.2 & 22.3 & 28.3 & 34.0 & 40.3
& 40.5 & 46.6 & 22.6 & 30.1 & 34.5 & 41.2
& 34.2 & 40.7 \\
Combo~\cite{fang2025combo}           & 39.2 & 45.7 & 21.6 & 29.9 & 33.3 & 40.0
& 39.7 & 46.2 & 21.5 & 31.1 & 33.6 & 41.2
& 33.4 & 40.6 \\
SimCIS~\cite{zhu2025rethinking}           & \textbf{42.4} & \textbf{56.7}
& 24.9 & \textbf{39.4}
& 36.5 & \textbf{51.1}
& \textbf{42.3} & 55.5
& 27.1 & 37.0
& 37.2 & 49.4
& 36.8 & 50.2 \\
\midrule

\textbf{FuTCR (Ours)}
& 42.0 & 56.1
& \textbf{28.0} & 38.0
& \textbf{37.3} & 50.3
& 42.1 & \textbf{58.2}
& \textbf{29.4} & \textbf{38.1}
& \textbf{37.9} & \textbf{51.7}
& \textbf{37.6} & \textbf{51.0} \\

\bottomrule
\end{tabular}%
\end{table}

\begin{figure*}[t]
    \centering
    \includegraphics[width=1.0\textwidth]{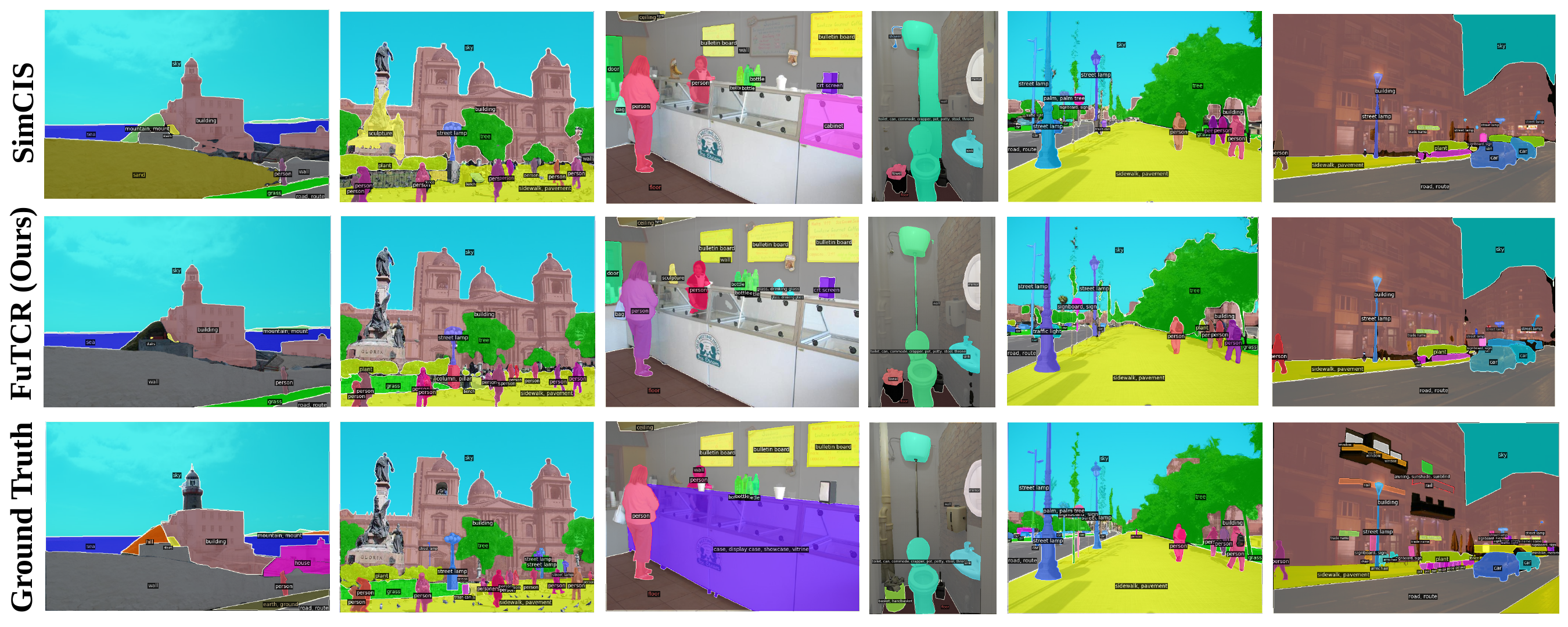}
    \caption{Qualitative comparison of segmentation produced by FuTCR and SimCIS~\cite{zhu2025rethinking} on ADE20K $100$–$50$. 
        FuTCR yields consistently richer panoptic reconstructions that recovers new-class regions or instances that SimCIS misses, while preserving cleaner, more complete segmentation of small and thin structures (\eg, wall, grass, bulletin board, etc.)
    }
     \label{fig:qualitative}
\end{figure*}

\cref{fig:qualitative} provides qualitative comparisons between SimCIS and FuTCR on the ADE20K $100$ - $50$ setting. SimCIS often suppresses newly introduced regions by assigning them to background or merging them with previously learned categories, producing fragmented or incomplete panoptic masks. In contrast, FuTCR yields more coherent masks for new classes while maintaining the structure of old-class regions. These examples support the quantitative trends in \cref{tab:main_results_100-5,tab:main_results_100-10,tab:main_results_100-50}, suggesting that future-targeted contrast and repulsion improve the model's ability to integrate newly labeled categories rather than only preserving past classes.

\begin{figure}[t]
  \centering
  \begin{minipage}[c]{0.57\linewidth}
    \centering
    \includegraphics[width=\linewidth]{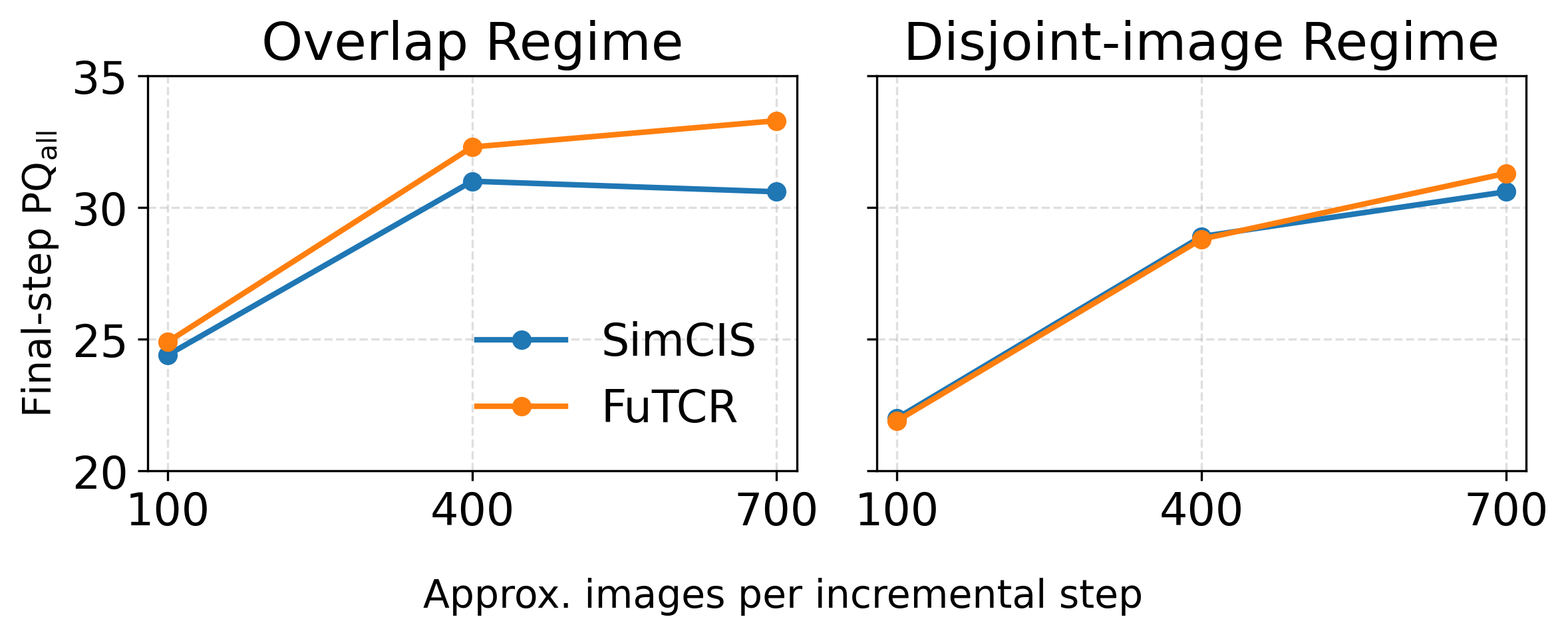}
  \end{minipage}%
  \hspace{2mm}
  \begin{minipage}[c]{0.40\linewidth}
    \caption{
      Robustness to reduced incremental supervision on ADE20K 100--5.
      We keep the same base step, class order, and ten incremental steps, and vary the average number of images per incremental step (x-axis: $\approx\!100$, $400$, and $700$).
      The left and right panels show final-step PQ$_\text{all}$ under the overlap and disjoint-image streams, respectively.
    }
    \label{fig:reduced_data_results}
  \end{minipage}
\end{figure}

Figure~\ref{fig:reduced_data_results} evaluates robustness to reduced incremental supervision on the 100--5 protocol. We keep the same base step, class order, and ten incremental steps, but progressively reduce the average number of training images per incremental step from roughly $700$ to $400$ and $100$. In the overlap stream, FuTCR consistently outperforms SimCIS across all supervision levels, with the largest relative gains at full data. In the disjoint-image stream, the methods are closer, though FuTCR still slightly improves PQ$_\text{all}$ in the full-data case, suggesting that structuring future-like regions is most beneficial when future-category evidence appears as unlabeled background and that extremely sparse supervision limits recoverable gains.

\noindent\textbf{Ablation Study.}
\begin{table}[tb]

\label{tab:ablation_results_100-5}
\centering

\caption{
Component ablation of FuTCR on the ADE20K 100--5 validation set.
We report final-step validation PQ on base classes (PQ$_\text{base}$ $1-100$), new classes
(PQ$_\text{new}$ $101-150$), and all classes (PQ$_\text{all}$ $1-150$) under overlap and disjoint-image streams. RC denotes region contrast and KFR denotes known-future repulsion.  We find both components are needed for best performance.
}
\label{tab:ablation_results}
\centering
\small
\begin{tabular}{lcc|ccc|ccc|c}
\toprule
\multirow{2}{*}{Variant} & \multirow{2}{*}{RC} & \multirow{2}{*}{KFR}
& \multicolumn{3}{c|}{Overlap Val.}
& \multicolumn{3}{c|}{Disjoint Val.}
& \multirow{2}{*}{Val. Avg.} \\

& & & 1-100 & 101-150 & all & 1-100 & 101-150 & all  & \\
\midrule
SimCIS~\cite{zhu2025rethinking}    & -- & -- & 37.7 & 19.3 & 31.5 & 35.1 & 22.6 & 30.9 &   31.2 \\
FuTCR-RC  & \checkmark & -- & 38.8 & 15.2 & 31.0 & 34.9 & 20.7 & 30.1 & 30.5  \\
FuTCR-KFR & -- & \checkmark & \textbf{39.1} & 20.2 & 32.9 & 35.7 & \textbf{23.4} & 31.6 &  32.2 \\
FuTCR     & \checkmark & \checkmark & 38.8  & \textbf{23.7} & \textbf{33.7} & \textbf{36.4} & 23.3 & \textbf{32.0} & \textbf{32.8} \\
\bottomrule
\end{tabular}
\end{table}

\cref{tab:ablation_results} quantifies the contributions of region contrast (RC) and known-future repulsion (KFR) on the ADE20K 100--5 \emph{validation} set. We select the final FuTCR configuration based on validation PQ$_\text{all}$ and \emph{do not} use test performance for model selection. KFR provides the strongest standalone improvement: relative to SimCIS, average validation PQ$_\text{all}$ increases from $31.2$ to $32.2$ (about $+3\%$), with gains in both overlap and disjoint streams. This matches our diagnosis that the main failure mode is future-class evidence collapsing into known-class decision regions, and that explicitly repelling future-like features from known-class prototypes directly mitigates this effect.

RC plays a complementary role. On its own, RC slightly reduces average PQ$_\text{all}$ (from $31.2$ to $30.5$), reflecting that organizing future-like regions into coherent prototypes is insufficient if those prototypes remain near known-class regions. When combined with KFR, however, RC boosts validation PQ$_\text{all}$ further from $32.2$ to $32.8$ (roughly an additional $+2\%$ relative), yielding the best overall validation performance and the configuration used in all main experiments. This pattern supports our design: future-aware region structuring is most effective when paired with an explicit separation objective that prevents future-like prototypes from being absorbed into old-class subspaces.

\subsection{Future-Aware Error and Representation Analysis}
\label{subsec:future_aware_analysis}

\begin{figure}[t]
  \begin{minipage}[c]{0.5\textwidth}
  \centering
    \includegraphics[width=0.9\linewidth]{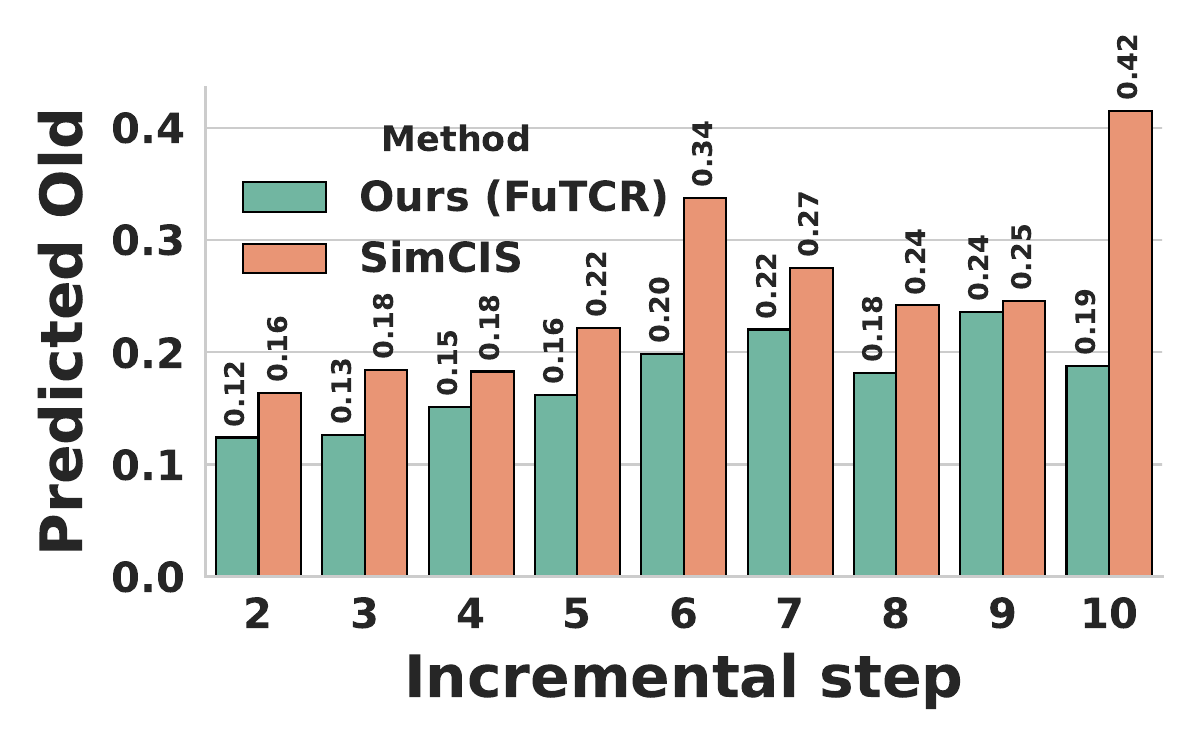}
  \end{minipage}
  \hspace{1mm}
  \begin{minipage}[c]{0.45\textwidth}
    \caption{
    \textbf{Future-aware error dynamics.} FuTCR reduces the fraction of future-class pixels that are misclassified as base classes across incremental steps compared to SimCIS~\cite{zhu2025rethinking}, indicating less future--old class confusion at the base step.
    }
    \label{fig:future_old_confusion_plot}
  \end{minipage}
\end{figure}

\textbf{Future-Category Confusion Profiling}%
\phantomsection\label{subsec:future_confusion_profiling}.
Motivated by the tendency of background-assigned future-class pixels to be absorbed into known-class decision regions during continual training, we explicitly quantify how base-step models allocate probability mass to pixels that will later belong to novel categories. For each incremental step, we compute, at the base step, the fraction of pixels whose ground-truth class lies in $\mathcal{C} \setminus \mathcal{C}^{\leq t}$ that are predicted as old classes, as background, or as future classes (when future-class logits are exposed). This defines a compact “future confusion profile’’ for each method, summarized in~\cref{fig:future_old_confusion_plot}.

Strong continual baselines already exhibit a marked tendency to fold future-category evidence into previously learned classes: in our experiments, the baseline misclassifies roughly $12$–$16\%$ of future pixels as old classes at early steps, and this future–old class confusion persists throughout training (see~\cref{fig:future_old_confusion_plot} and~\cite{fang2025combo,zhu2025rethinking,lin2023preparing}). FuTCR consistently reduces this fraction across all steps while increasing the share of future pixels that remain separable from old-class decision regions, providing quantitative evidence that our future-aware objectives reshape the previous-step feature space in a way that is more conducive to integrating new categories.

\begin{figure*}[t]
    \centering
    \begin{minipage}[t]{0.645\textwidth}
        \centering
        \includegraphics[width=\textwidth]{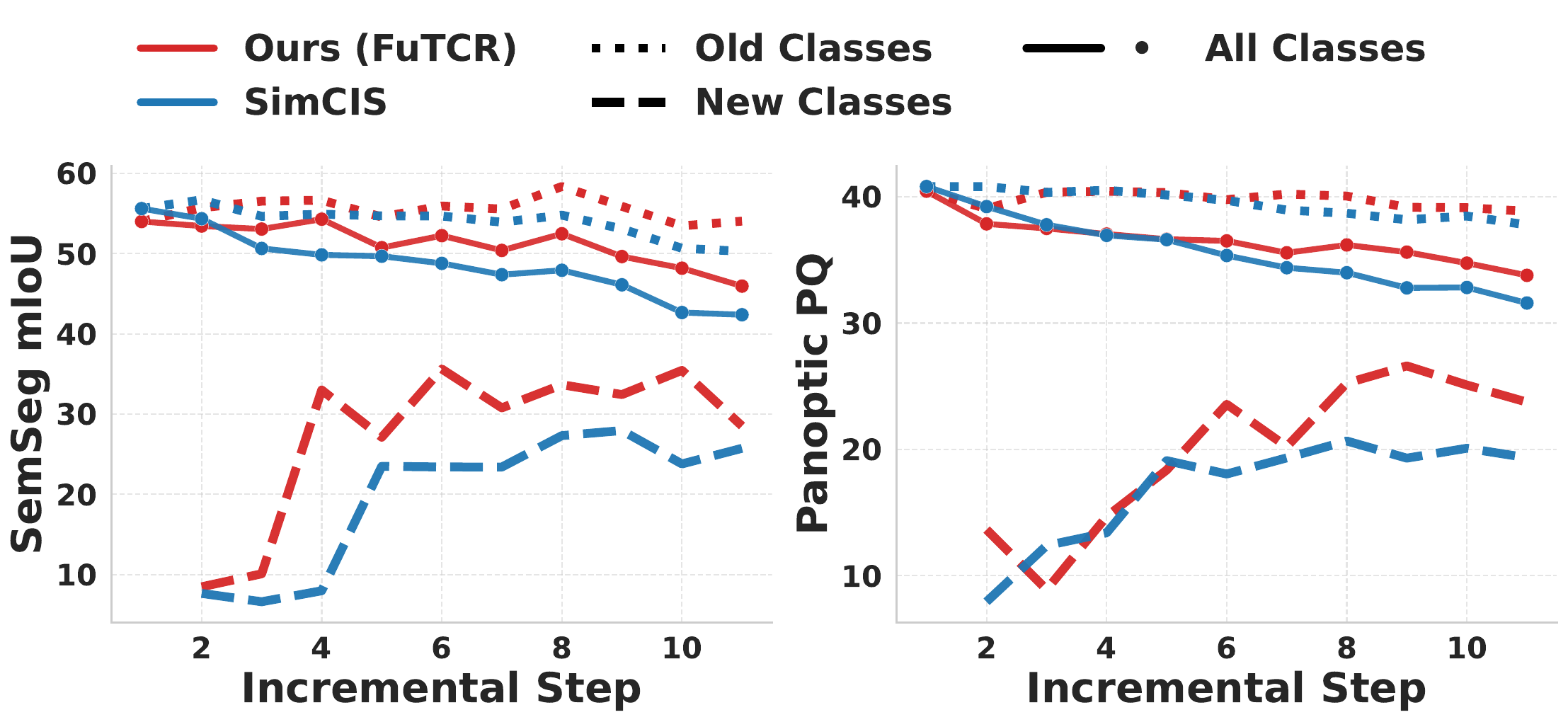}
    \end{minipage}\hfill
    \rule{0.05pt}{0.17\textheight}%
    \hfill
    \begin{minipage}[t]{0.345\textwidth}
        \centering
        \includegraphics[width=\textwidth]{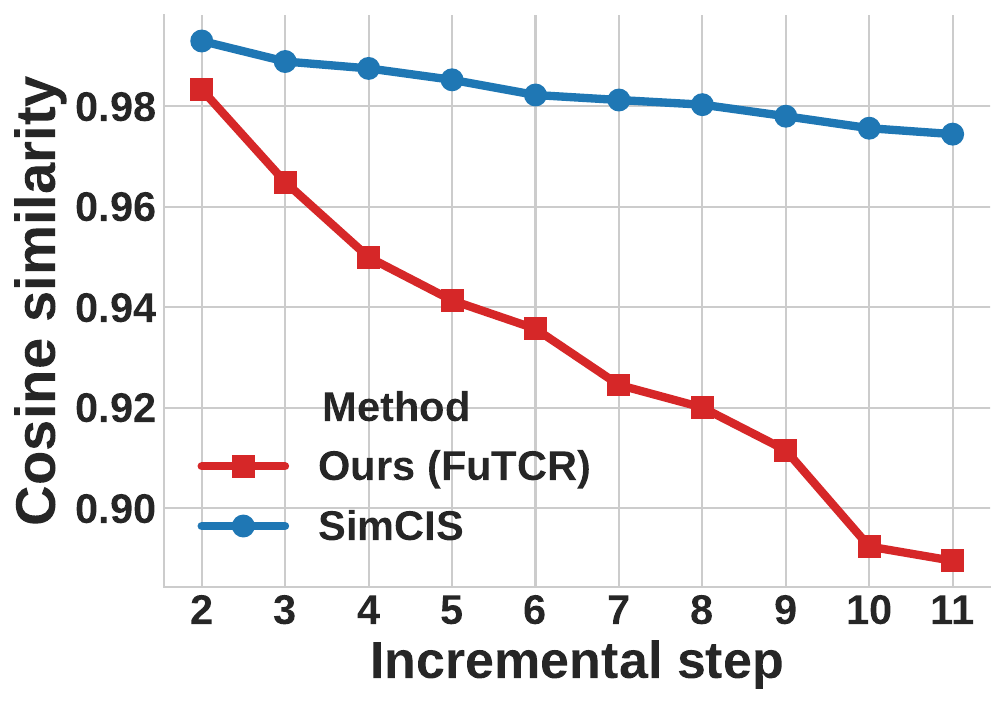}
    \end{minipage}
    \caption{
    \textbf{Left/middle}: FuTCR consistently outperforms SimCIS~\cite{zhu2025rethinking} in mIoU and PQ on ADE20K 100--5, with especially pronounced gains on newly introduced classes and improved performance on previously learned classes. \textbf{Right}: cross-step prototype similarity for old classes, where FuTCR permits moderate drift (down to $\approx 0.89$) instead of the baseline’s near-rigid prototypes ($>0.97$), and this adaptive evolution coincides with higher PQ and reduced future--old class confusion.
    }
    \label{fig:incre_miou_pq_and_feat_sim}
\end{figure*}

\textbf{Cross-Step Prototype Congruence}%
\phantomsection\label{subsec:feature_congruity}
Prototype-based continual methods are often evaluated by how well they preserve old-class feature centroids across steps, e.g., via cosine similarity between current and initial prototypes~\cite{cermelli2023comformer,lin2023preparing}. However, our measurements show that such nearly frozen prototypes do not reliably correlate with better future-class behavior. In the baseline, mean cosine similarity between current and step-1 prototypes remains above $0.97$ for all steps~\cref{fig:incre_miou_pq_and_feat_sim}(right), yet the model still exhibits strong future–old class confusion.

FuTCR instead allows old-class prototypes to evolve more over time: mean cosine similarity gradually decreases from $0.98$ at step~2 to about $0.89$ by step~11~\cref{fig:incre_miou_pq_and_feat_sim}(right), while panoptic quality improves and future–old class confusion decreases~\cref{fig:future_old_confusion_plot} and~\cref{tab:main_results_100-5}. This pattern indicates that a controlled degree of adaptive prototype drift, rather than strict preservation of initial centroids, is beneficial for carving out a dedicated subspace for unseen categories and for stabilizing future-class integration in continual panoptic segmentation.

\section{Conclusion}
\label{sec:conclusion_dis}
We introduced FuTCR, a future-targeted contrastive and repulsive learning framework for continual panoptic segmentation. FuTCR directly addresses the tendency of standard protocols to absorb unlabeled future-class regions into background or known-class decision regions by discovering future-like regions from model predictions, organizing them via pixel-to-region contrast, and repelling them from known-class prototypes. Across controlled overlap and disjoint-image streams on ADE20K, FuTCR improves new-class integration while maintaining competitive base-class performance, with the clearest gains in the long 100--5 protocol and consistent improvements in shorter and reduced-data settings. These results indicate that background regions contain exploitable structure for preparing representations before future class labels arrive.

\noindent\textbf{Limitations and Future Work.}
Our study is restricted to ADE20K, a specific set of overlap regimes, and a small number of random seeds and architectures, so even broader robustness remains to be explored. Extending FuTCR to other continual segmentation settings and task orderings, and refining future-like region selection, are important next steps.

\bibliographystyle{unsrtnat} %
\bibliography{main} 

\begin{thebibliography}{60}
\providecommand{\natexlab}[1]{#1}
\providecommand{\url}[1]{\texttt{#1}}
\expandafter\ifx\csname urlstyle\endcsname\relax
  \providecommand{\doi}[1]{doi: #1}\else
  \providecommand{\doi}{doi: \begingroup \urlstyle{rm}\Url}\fi

\bibitem[Zhou et~al.(2025)Zhou, Tian, Lv, Shi, Li, Ye, Zhang, and
  Lv]{zhou2025ferret}
Yuhao Zhou, Yuxin Tian, Jindi Lv, Mingjia Shi, Yuanxi Li, Qing Ye, Shuhao
  Zhang, and Jiancheng Lv.
\newblock Ferret: An efficient online continual learning framework under
  varying memory constraints.
\newblock In \emph{Proceedings of the Computer Vision and Pattern Recognition
  Conference}, pages 4850--4861, 2025.

\bibitem[Kang et~al.(2025)Kang, Seifer, Lee, and Ryu]{kang2025your}
Hankyul Kang, Gregor Seifer, Donghyun Lee, and Jongbin Ryu.
\newblock Do your best and get enough rest for continual learning.
\newblock In \emph{Proceedings of the Computer Vision and Pattern Recognition
  Conference}, pages 10077--10086, 2025.

\bibitem[Castro et~al.(2018)Castro, Mar{\'\i}n-Jim{\'e}nez, Guil, Schmid, and
  Alahari]{castro2018end}
Francisco~M Castro, Manuel~J Mar{\'\i}n-Jim{\'e}nez, Nicol{\'a}s Guil, Cordelia
  Schmid, and Karteek Alahari.
\newblock End-to-end incremental learning.
\newblock In \emph{Proceedings of the European conference on computer vision
  (ECCV)}, pages 233--248, 2018.

\bibitem[Rebuffi et~al.(2017)Rebuffi, Kolesnikov, Sperl, and
  Lampert]{rebuffi2017icarl}
Sylvestre-Alvise Rebuffi, Alexander Kolesnikov, Georg Sperl, and Christoph~H
  Lampert.
\newblock icarl: Incremental classifier and representation learning.
\newblock In \emph{Proceedings of the IEEE conference on Computer Vision and
  Pattern Recognition}, pages 2001--2010, 2017.

\bibitem[Chaudhry et~al.(2018)Chaudhry, Dokania, Ajanthan, and
  Torr]{chaudhry2018riemannian}
Arslan Chaudhry, Puneet~K Dokania, Thalaiyasingam Ajanthan, and Philip~HS Torr.
\newblock Riemannian walk for incremental learning: Understanding forgetting
  and intransigence.
\newblock In \emph{Proceedings of the European conference on computer vision
  (ECCV)}, pages 532--547, 2018.

\bibitem[Shin et~al.(2017)Shin, Lee, Kim, and Kim]{shin2017continual}
Hanul Shin, Jung~Kwon Lee, Jaehong Kim, and Jiwon Kim.
\newblock Continual learning with deep generative replay.
\newblock \emph{Advances in neural information processing systems}, 30, 2017.

\bibitem[Yan et~al.(2021)Yan, Xie, and He]{yan2021dynamically}
Shipeng Yan, Jiangwei Xie, and Xuming He.
\newblock Der: Dynamically expandable representation for class incremental
  learning.
\newblock In \emph{Proceedings of the IEEE/CVF conference on computer vision
  and pattern recognition}, pages 3014--3023, 2021.

\bibitem[Tao et~al.(2020)Tao, Hong, Chang, Dong, Wei, and Gong]{tao2020few}
Xiaoyu Tao, Xiaopeng Hong, Xinyuan Chang, Songlin Dong, Xing Wei, and Yihong
  Gong.
\newblock Few-shot class-incremental learning.
\newblock In \emph{Proceedings of the IEEE/CVF conference on computer vision
  and pattern recognition}, pages 12183--12192, 2020.

\bibitem[Jiao et~al.(2024)Jiao, Lai, Li, and Xu]{jiao2024vector}
Li~Jiao, Qiuxia Lai, Yu~Li, and Qiang Xu.
\newblock Vector quantization prompting for continual learning.
\newblock \emph{Advances in Neural Information Processing Systems},
  37:\penalty0 34056--34076, 2024.

\bibitem[Wang et~al.(2021)Wang, Yang, Li, Hong, Li, and Zhu]{wang2021ordisco}
Liyuan Wang, Kuo Yang, Chongxuan Li, Lanqing Hong, Zhenguo Li, and Jun Zhu.
\newblock Ordisco: Effective and efficient usage of incremental unlabeled data
  for semi-supervised continual learning.
\newblock In \emph{Proceedings of the IEEE/CVF Conference on Computer Vision
  and Pattern Recognition}, pages 5383--5392, 2021.

\bibitem[Friedman and Meir(2026)]{friedman2026pacbayes}
Lior Friedman and Ron Meir.
\newblock {PAC}-bayes bounds for cumulative loss in continual learning.
\newblock In \emph{The Fourteenth International Conference on Learning
  Representations}, 2026.
\newblock URL \url{https://openreview.net/forum?id=hWw269fPov}.

\bibitem[Ye et~al.(2025)Ye, Zhao, Liu, Chen, Bors, Sun, Hu, and
  Zhou]{ye2025dynamic}
Fei Ye, Yulong Zhao, Qihe Liu, Junlin Chen, Adrian~G. Bors, Jingling Sun,
  Rongyao Hu, and Shijie Zhou.
\newblock Dynamic siamese expansion framework for improving robustness in
  online continual learning.
\newblock In \emph{Advances in Neural Information Processing Systems}, 2025.

\bibitem[Wu et~al.(2025)Wu, Wang, Chen, Zhang, Tan, Liu, and
  Li]{wu2025exploiting}
Yanru Wu, Jianning Wang, Xiangyu Chen, Enming Zhang, Yang Tan, Hanbing Liu, and
  Yang Li.
\newblock Exploiting task relationships in continual learning via
  transferability-aware task embeddings.
\newblock In \emph{Advances in Neural Information Processing Systems}, 2025.
\newblock URL \url{https://openreview.net/forum?id=V8FnYzDX35}.

\bibitem[Momeni et~al.(2025)Momeni, Xiao, and Liu]{momeni2025anacp}
Saleh Momeni, Changnan Xiao, and Bing Liu.
\newblock Anacp: Toward upper‐bound continual learning via analytic
  contrastive projection.
\newblock In \emph{Advances in Neural Information Processing Systems}, 2025.
\newblock URL \url{https://openreview.net/forum?id=qQbvLU34F1}.

\bibitem[chenyanxi et~al.(2026)chenyanxi, Li, Yuyang, Wang, Li, Li, Li, Wei,
  and Wu]{chenyanxi2026hippotune}
chenyanxi, Xiuxing Li, Han Yuyang, Zhuo Wang, Qing Li, Ziyu Li, Xiang Li, Chen
  Wei, and Xia Wu.
\newblock Hippotune: A hippocampal associative loop{\textendash}inspired
  fine-tuning method for continual learning.
\newblock In \emph{The Fourteenth International Conference on Learning
  Representations}, 2026.
\newblock URL \url{https://openreview.net/forum?id=MtDiLnnYgm}.

\bibitem[Yang et~al.(2025)Yang, Dong, Li, Song, and Lin]{yang2025adapt}
Ze~Yang, Shichao Dong, Ruibo Li, Nan Song, and Guosheng Lin.
\newblock {ADAPT}: Attentive self-distillation and dual-decoder prediction
  fusion for continual panoptic segmentation.
\newblock In \emph{The Thirteenth International Conference on Learning
  Representations}, 2025.
\newblock URL \url{https://openreview.net/forum?id=HF1UmIVv6a}.

\bibitem[Yin et~al.(2025)Yin, Feng, Lyu, Shang, Liu, Feng, and
  Wan]{yin2025beyond}
Hongmei Yin, Tingliang Feng, Fan Lyu, Fanhua Shang, Hongying Liu, Wei Feng, and
  Liang Wan.
\newblock Beyond background shift: Rethinking instance replay in continual
  semantic segmentation.
\newblock In \emph{Proceedings of the Computer Vision and Pattern Recognition
  Conference}, pages 9839--9848, 2025.

\bibitem[Truong et~al.(2025)Truong, Prabhu, Raj, Cothren, and
  Luu]{truong2025falcon}
Thanh-Dat Truong, Utsav Prabhu, Bhiksha Raj, Jackson Cothren, and Khoa Luu.
\newblock Falcon: Fairness learning via contrastive attention approach to
  continual semantic scene understanding.
\newblock In \emph{Proceedings of the Computer Vision and Pattern Recognition
  Conference}, pages 15065--15075, 2025.

\bibitem[Zhu et~al.(2025{\natexlab{a}})Zhu, Wang, Shao, dong Yang, Sang, and
  Gao]{zhu2025continual}
Guilin Zhu, Runmin Wang, Yuanjie Shao, Wei dong Yang, Nong Sang, and Changxin
  Gao.
\newblock Continual gaussian mixture distribution modeling for class
  incremental semantic segmentation.
\newblock In \emph{The Thirty-ninth Annual Conference on Neural Information
  Processing Systems}, 2025{\natexlab{a}}.
\newblock URL \url{https://openreview.net/forum?id=dtYKDOBkc7}.

\bibitem[Zhang et~al.(2025)Zhang, Zou, Chen, Zhou, Wang, Zhong, and
  Yan]{zhang2025parameter}
Xinyue Zhang, Xu~Zou, Liqun Chen, Jiahuan Zhou, Guodong Wang, Sheng Zhong, and
  Luxin Yan.
\newblock Parameter release and knowledge reuse for class-incremental semantic
  segmentation, 2025.
\newblock URL \url{https://openreview.net/forum?id=9qbKOaF8YJ}.

\bibitem[Fang et~al.(2025)Fang, Zhang, Gao, Jiao, Liu, and Wei]{fang2025combo}
Kai Fang, Anqi Zhang, Guangyu Gao, Jianbo Jiao, Chi~Harold Liu, and Yunchao
  Wei.
\newblock Combo: Conflict mitigation via branched optimization for class
  incremental segmentation.
\newblock In \emph{Proceedings of the Computer Vision and Pattern Recognition
  Conference}, pages 25667--25676, 2025.

\bibitem[Zhu et~al.(2023)Zhu, Chen, Yin, See, and Liu]{zhu2023continual}
Lanyun Zhu, Tianrun Chen, Jianxiong Yin, Simon See, and Jun Liu.
\newblock Continual semantic segmentation with automatic memory sample
  selection.
\newblock In \emph{Proceedings of the IEEE/CVF Conference on Computer Vision
  and Pattern Recognition}, pages 3082--3092, 2023.

\bibitem[Zhu et~al.(2025{\natexlab{b}})Zhu, Shi, Wang, Tang, Wei, Wu, Li, and
  Yang]{zhu2025rethinking}
Yuchen Zhu, Cheng Shi, Dingyou Wang, Jiajin Tang, Zhengxuan Wei, Yu~Wu, Guanbin
  Li, and Sibei Yang.
\newblock Rethinking query-based transformer for continual image segmentation.
\newblock In \emph{Proceedings of the Computer Vision and Pattern Recognition
  Conference}, pages 4595--4606, 2025{\natexlab{b}}.

\bibitem[Tang et~al.(2023)Tang, Zheng, Shi, and Yang]{tang2023contrastive}
Jiajin Tang, Ge~Zheng, Cheng Shi, and Sibei Yang.
\newblock Contrastive grouping with transformer for referring image
  segmentation.
\newblock In \emph{Proceedings of the IEEE/CVF conference on computer vision
  and pattern recognition}, pages 23570--23580, 2023.

\bibitem[Shi and Yang(2024)]{shi2024devil}
Cheng Shi and Sibei Yang.
\newblock The devil is in the object boundary: Towards annotation-free instance
  segmentation using foundation models.
\newblock \emph{arXiv preprint arXiv:2404.11957}, 2024.

\bibitem[Zhang et~al.(2023)Zhang, Gao, Jiao, Liu, and Wei]{zhang2023coinseg}
Zekang Zhang, Guangyu Gao, Jianbo Jiao, Chi~Harold Liu, and Yunchao Wei.
\newblock Coinseg: Contrast inter-and intra-class representations for
  incremental segmentation.
\newblock In \emph{Proceedings of the IEEE/CVF International Conference on
  Computer Vision}, pages 843--853, 2023.

\bibitem[Lin et~al.(2025)Lin, Wang, and Wang]{Lin_2025_CVPR_universal_segmen}
Zihan Lin, Zilei Wang, and Xu~Wang.
\newblock Towards continual universal segmentation.
\newblock In \emph{Proceedings of the IEEE/CVF Conference on Computer Vision
  and Pattern Recognition (CVPR)}, pages 29417--29427, June 2025.

\bibitem[Kim et~al.(2024)Kim, Yu, and Hwang]{kim2024eclipse}
Beomyoung Kim, Joonsang Yu, and Sung~Ju Hwang.
\newblock Eclipse: Efficient continual learning in panoptic segmentation with
  visual prompt tuning.
\newblock In \emph{Proceedings of the IEEE/CVF Conference on Computer Vision
  and Pattern Recognition}, pages 3346--3356, 2024.

\bibitem[Cermelli et~al.(2023)Cermelli, Cord, and
  Douillard]{cermelli2023comformer}
Fabio Cermelli, Matthieu Cord, and Arthur Douillard.
\newblock Comformer: Continual learning in semantic and panoptic segmentation.
\newblock In \emph{Proceedings of the IEEE/CVF Conference on Computer Vision
  and Pattern Recognition}, pages 3010--3020, 2023.

\bibitem[Chen et~al.(2024)Chen, Cong, Luo, Ip, and Kwong]{chen2024strike}
Jinpeng Chen, Runmin Cong, Yuxuan Luo, Horace Ho~Shing Ip, and Sam Kwong.
\newblock Strike a balance in continual panoptic segmentation.
\newblock In \emph{European Conference on Computer Vision}, pages 126--142.
  Springer, 2024.

\bibitem[Sohn et~al.(2020)Sohn, Berthelot, Carlini, Zhang, Zhang, Raffel,
  Cubuk, Kurakin, and Li]{sohn2020fixmatch}
Kihyuk Sohn, David Berthelot, Nicholas Carlini, Zizhao Zhang, Han Zhang,
  Colin~A Raffel, Ekin~Dogus Cubuk, Alexey Kurakin, and Chun-Liang Li.
\newblock Fixmatch: Simplifying semi-supervised learning with consistency and
  confidence.
\newblock \emph{Advances in neural information processing systems},
  33:\penalty0 596--608, 2020.

\bibitem[Chen et~al.(2020)Chen, Kornblith, Norouzi, and Hinton]{chen2020simple}
Ting Chen, Simon Kornblith, Mohammad Norouzi, and Geoffrey Hinton.
\newblock A simple framework for contrastive learning of visual
  representations.
\newblock In \emph{International conference on machine learning}, pages
  1597--1607. PmLR, 2020.

\bibitem[He et~al.(2020)He, Fan, Wu, Xie, and Girshick]{he2020momentum}
Kaiming He, Haoqi Fan, Yuxin Wu, Saining Xie, and Ross Girshick.
\newblock Momentum contrast for unsupervised visual representation learning.
\newblock In \emph{Proceedings of the IEEE/CVF conference on computer vision
  and pattern recognition}, pages 9729--9738, 2020.

\bibitem[Olsson et~al.(2021)Olsson, Tranheden, Pinto, and
  Svensson]{olsson2021classmix}
Viktor Olsson, Wilhelm Tranheden, Juliano Pinto, and Lennart Svensson.
\newblock Classmix: Segmentation-based data augmentation for semi-supervised
  learning.
\newblock In \emph{Proceedings of the IEEE/CVF winter conference on
  applications of computer vision}, pages 1369--1378, 2021.

\bibitem[Lin et~al.(2023)Lin, Wang, and Zhang]{lin2023preparing}
Zihan Lin, Zilei Wang, and Yixin Zhang.
\newblock Preparing the future for continual semantic segmentation.
\newblock In \emph{Proceedings of the IEEE/CVF International Conference on
  Computer Vision}, pages 11910--11920, 2023.

\bibitem[Yuan et~al.(2024)Yuan, Zhao, Liu, Li, and Li]{yuan2024continual}
Bo~Yuan, Danpei Zhao, Zhuoran Liu, Wentao Li, and Tian Li.
\newblock Continual panoptic perception: Towards multi-modal incremental
  interpretation of remote sensing images.
\newblock In \emph{Proceedings of the 32nd ACM International Conference on
  Multimedia}, pages 2117--2126, 2024.

\bibitem[Cha et~al.(2021)Cha, Yoo, Moon, et~al.]{cha2021ssul}
Sungmin Cha, YoungJoon Yoo, Taesup Moon, et~al.
\newblock Ssul: Semantic segmentation with unknown label for exemplar-based
  class-incremental learning.
\newblock \emph{Advances in neural information processing systems},
  34:\penalty0 10919--10930, 2021.

\bibitem[Manjunath et~al.(2025)Manjunath, Madhu, Sikdar, and
  Sundaram]{manjunath2025vista}
D~Manjunath, Shrikar Madhu, Aniruddh Sikdar, and Suresh Sundaram.
\newblock Vista-clip: Visual incremental self-tuned adaptation for efficient
  continual panoptic segmentation.
\newblock In \emph{2025 IEEE/CVF Conference on Computer Vision and Pattern
  Recognition Workshops (CVPRW)}, pages 6557--6565. IEEE, 2025.

\bibitem[French(1999)]{french1999catastrophic}
Robert~M French.
\newblock Catastrophic forgetting in connectionist networks.
\newblock \emph{Trends in cognitive sciences}, 3\penalty0 (4):\penalty0
  128--135, 1999.

\bibitem[Robins(1995)]{robins1995catastrophic}
Anthony Robins.
\newblock Catastrophic forgetting, rehearsal and pseudorehearsal.
\newblock \emph{Connection Science}, 7\penalty0 (2):\penalty0 123--146, 1995.

\bibitem[Thrun(1998)]{thrun1998lifelong}
Sebastian Thrun.
\newblock Lifelong learning algorithms.
\newblock In \emph{Learning to learn}, pages 181--209. Springer, 1998.

\bibitem[Yuan and Zhao(2024)]{yuan2024survey}
Bo~Yuan and Danpei Zhao.
\newblock A survey on continual semantic segmentation: Theory, challenge,
  method and application.
\newblock \emph{IEEE Transactions on Pattern Analysis and Machine
  Intelligence}, 2024.

\bibitem[Park et~al.(2024)Park, Moon, Lee, Kim, and Heo]{park2024mitigating}
Gilhan Park, WonJun Moon, SuBeen Lee, Tae-Young Kim, and Jae-Pil Heo.
\newblock Mitigating background shift in class-incremental semantic
  segmentation.
\newblock In \emph{European Conference on Computer Vision}, pages 71--88.
  Springer, 2024.

\bibitem[Cermelli et~al.(2020)Cermelli, Mancini, Bulo, Ricci, and
  Caputo]{cermelli2020modelingMIB}
Fabio Cermelli, Massimiliano Mancini, Samuel~Rota Bulo, Elisa Ricci, and
  Barbara Caputo.
\newblock Modeling the background for incremental learning in semantic
  segmentation.
\newblock In \emph{Proceedings of the IEEE/CVF conference on computer vision
  and pattern recognition}, pages 9233--9242, 2020.

\bibitem[Kalb(2024)]{kalb2024principles}
Tobias~Michael Kalb.
\newblock \emph{Principles of Catastrophic Forgetting for Continual Semantic
  Segmentation in Automated Driving}.
\newblock KIT Scientific Publishing, 2024.

\bibitem[Lin et~al.(2022)Lin, Wang, and Zhang]{lin2022continual}
Zihan Lin, Zilei Wang, and Yixin Zhang.
\newblock Continual semantic segmentation via structure preserving and
  projected feature alignment.
\newblock In \emph{European Conference on Computer Vision}, pages 345--361.
  Springer, 2022.

\bibitem[Song et~al.(2023)Song, Zhang, and Shi]{song2023scale}
Zichen Song, Xiaoliang Zhang, and Zhaofeng Shi.
\newblock Scale-hybrid group distillation with knowledge disentangling for
  continual semantic segmentation.
\newblock \emph{Sensors}, 23\penalty0 (18):\penalty0 7820, 2023.

\bibitem[ElAraby et~al.(2024)ElAraby, Harakeh, and Paull]{elaraby2024bacs}
Mostafa ElAraby, Ali Harakeh, and Liam Paull.
\newblock Bacs: Background aware continual semantic segmentation.
\newblock \emph{arXiv preprint arXiv:2404.13148}, 2024.

\bibitem[Luo et~al.(2025)Luo, Chen, Cong, Ip, and Kwong]{luo2025trace}
Yuxuan Luo, Jinpeng Chen, Runmin Cong, Horace Ho~Shing Ip, and Sam Kwong.
\newblock Trace back and go ahead: Completing partial annotation for continual
  semantic segmentation.
\newblock \emph{Pattern Recognition}, 165:\penalty0 111613, 2025.

\bibitem[Hwang et~al.(2021)Hwang, Oh, Lee, and Han]{hwang2021exemplar}
Jaedong Hwang, Seoung~Wug Oh, Joon-Young Lee, and Bohyung Han.
\newblock Exemplar-based open-set panoptic segmentation network.
\newblock In \emph{Proceedings of the IEEE/CVF Conference on Computer Vision
  and Pattern Recognition}, pages 1175--1184, 2021.

\bibitem[Yin et~al.(2024)Yin, Chen, Zhou, Deng, Xu, and Li]{yin2024revisiting}
Yufei Yin, Hao Chen, Wengang Zhou, Jiajun Deng, Haiming Xu, and Houqiang Li.
\newblock Revisiting open-set panoptic segmentation.
\newblock In \emph{Proceedings of the AAAI Conference on Artificial
  Intelligence}, volume~38, pages 6747--6754, 2024.

\bibitem[Dong et~al.(2022)Dong, Chen, Yuan, Xie, Zhao, Yu, Dong, and
  Zhang]{dong2022region}
Hexin Dong, Zifan Chen, Mingze Yuan, Yutong Xie, Jie Zhao, Fei Yu, Bin Dong,
  and Li~Zhang.
\newblock Region-aware metric learning for open world semantic segmentation via
  meta-channel aggregation.
\newblock \emph{arXiv preprint arXiv:2205.08083}, 2022.

\bibitem[Xu et~al.(2022)Xu, Chen, Liu, and Yin]{xu2022dual}
Hai-Ming Xu, Hao Chen, Lingqiao Liu, and Yufei Yin.
\newblock Dual decision improves open-set panoptic segmentation.
\newblock \emph{arXiv preprint arXiv:2207.02504}, 2022.

\bibitem[Kirillov et~al.(2019)Kirillov, He, Girshick, Rother, and
  Doll{\'a}r]{kirillov2019panoptic}
Alexander Kirillov, Kaiming He, Ross Girshick, Carsten Rother, and Piotr
  Doll{\'a}r.
\newblock Panoptic segmentation.
\newblock In \emph{Proceedings of the IEEE/CVF conference on computer vision
  and pattern recognition}, pages 9404--9413, 2019.

\bibitem[Zheng et~al.(2021)Zheng, Chen, Yao, Yang, Li, Zhang, Zhang, Tsang,
  Zhou, and Zhou]{zheng2021contrastive}
Huangjie Zheng, Xu~Chen, Jiangchao Yao, Hongxia Yang, Chunyuan Li, Ya~Zhang,
  Hao Zhang, Ivor Tsang, Jingren Zhou, and Mingyuan Zhou.
\newblock Contrastive attraction and contrastive repulsion for representation
  learning.
\newblock \emph{arXiv preprint arXiv:2105.03746}, 2021.

\bibitem[B{\"o}hm et~al.(2022)B{\"o}hm, Berens, and Kobak]{bohm2022attraction}
Jan~Niklas B{\"o}hm, Philipp Berens, and Dmitry Kobak.
\newblock Attraction-repulsion spectrum in neighbor embeddings.
\newblock \emph{Journal of Machine Learning Research}, 23\penalty0
  (95):\penalty0 1--32, 2022.

\bibitem[Yu and Shi(2001)]{yu2001segmentation}
Stella~X. Yu and Jianbo Shi.
\newblock Segmentation with pairwise attraction and repulsion.
\newblock In \emph{Proceedings Eighth IEEE International Conference on Computer
  Vision. ICCV 2001}, volume~1, pages 52--58. IEEE, 2001.

\bibitem[Zhou et~al.(2017)Zhou, Zhao, Puig, Fidler, Barriuso, and
  Torralba]{zhou2017sceneade20k}
Bolei Zhou, Hang Zhao, Xavier Puig, Sanja Fidler, Adela Barriuso, and Antonio
  Torralba.
\newblock Scene parsing through ade20k dataset.
\newblock In \emph{Proceedings of the IEEE Conference on Computer Vision and
  Pattern Recognition}, 2017.

\bibitem[Cheng et~al.(2022)Cheng, Misra, Schwing, Kirillov, and
  Girdhar]{cheng2022masked}
Bowen Cheng, Ishan Misra, Alexander~G Schwing, Alexander Kirillov, and Rohit
  Girdhar.
\newblock Masked-attention mask transformer for universal image segmentation.
\newblock In \emph{Proceedings of the IEEE/CVF conference on computer vision
  and pattern recognition}, pages 1290--1299, 2022.

\bibitem[He et~al.(2016)He, Zhang, Ren, and Sun]{he2016deepresnet}
Kaiming He, Xiangyu Zhang, Shaoqing Ren, and Jian Sun.
\newblock Deep residual learning for image recognition.
\newblock In \emph{Proceedings of the IEEE conference on computer vision and
  pattern recognition}, pages 770--778, 2016.

\end{thebibliography}


\appendix

\newpage

\section{Additional Analysis}
\label{app:additional_analysis}

\subsection{Stability--plasticity Trajectory}
\label{app:stability_plasticity}
\begin{figure*}[ht]
    \centering
    \vspace{-10pt}
    \includegraphics[width=0.5\textwidth]{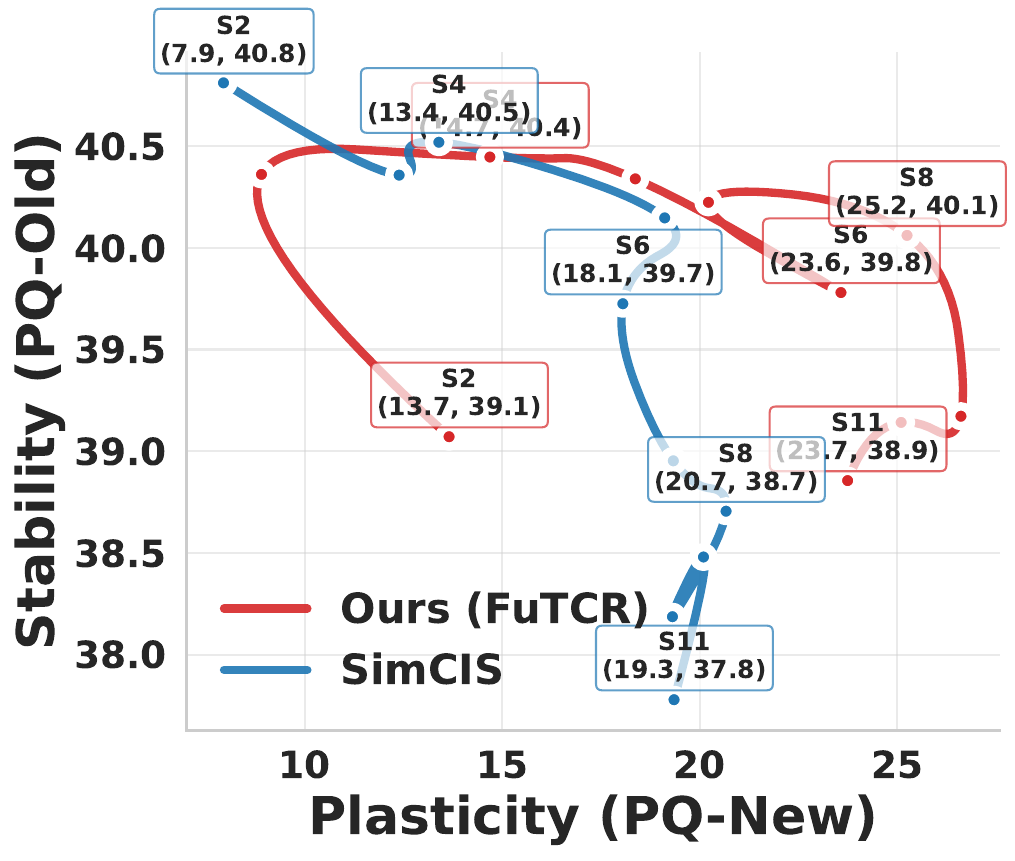}
    \caption{
        Stability--plasticity trajectory of FuTCR versus SimCIS on ADE20K 100--5 (steps 2--11). Each marker reflects the trade-off between base-class retention and new-class performance at a given step, with FuTCR occupying a more favorable region of the plane across the sequence.
    }
    \label{fig:stability_plasticity_plot}
    \vspace{-10pt}
\end{figure*}

\cref{fig:stability_plasticity_plot} summarizes the stability--plasticity behavior of FuTCR and SimCIS on ADE20K 100--5. Each point corresponds to a step $t \in \{2,\dots,11\}$, with the horizontal axis measuring base-class retention and the vertical axis measuring new-class performance. FuTCR traces a consistently more favorable trajectory, achieving stronger gains on new classes while incurring less degradation on previously learned classes, indicating that future-aware region structuring improves the long-term stability--plasticity trade-off in the most challenging incremental setting.

\subsection{Extra Qualitative Results}
\label{app:extra_qualitative_results}

\begin{figure*}[t]
    \centering
    \includegraphics[width=\textwidth]{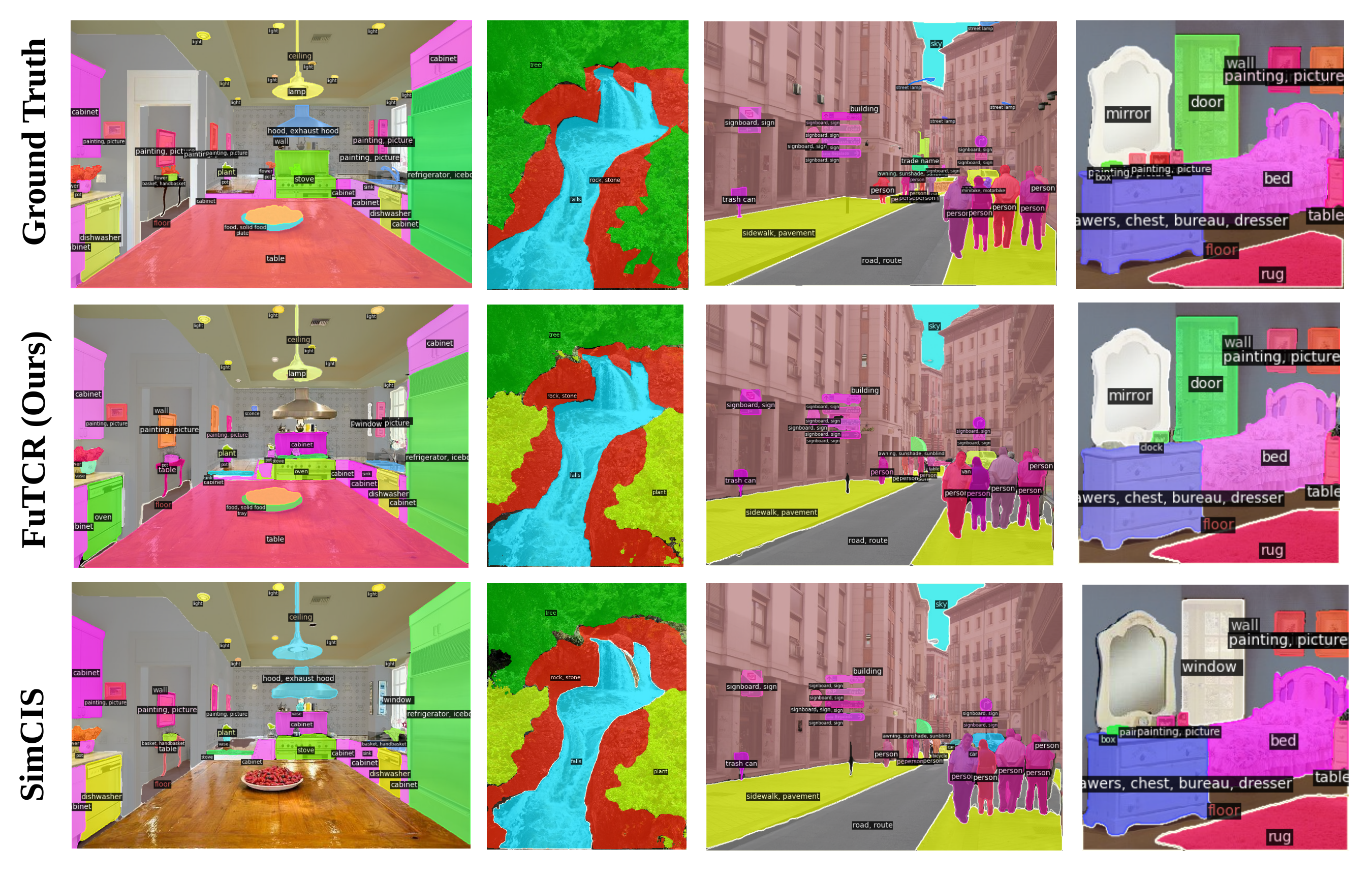}

    \includegraphics[width=0.98\textwidth]{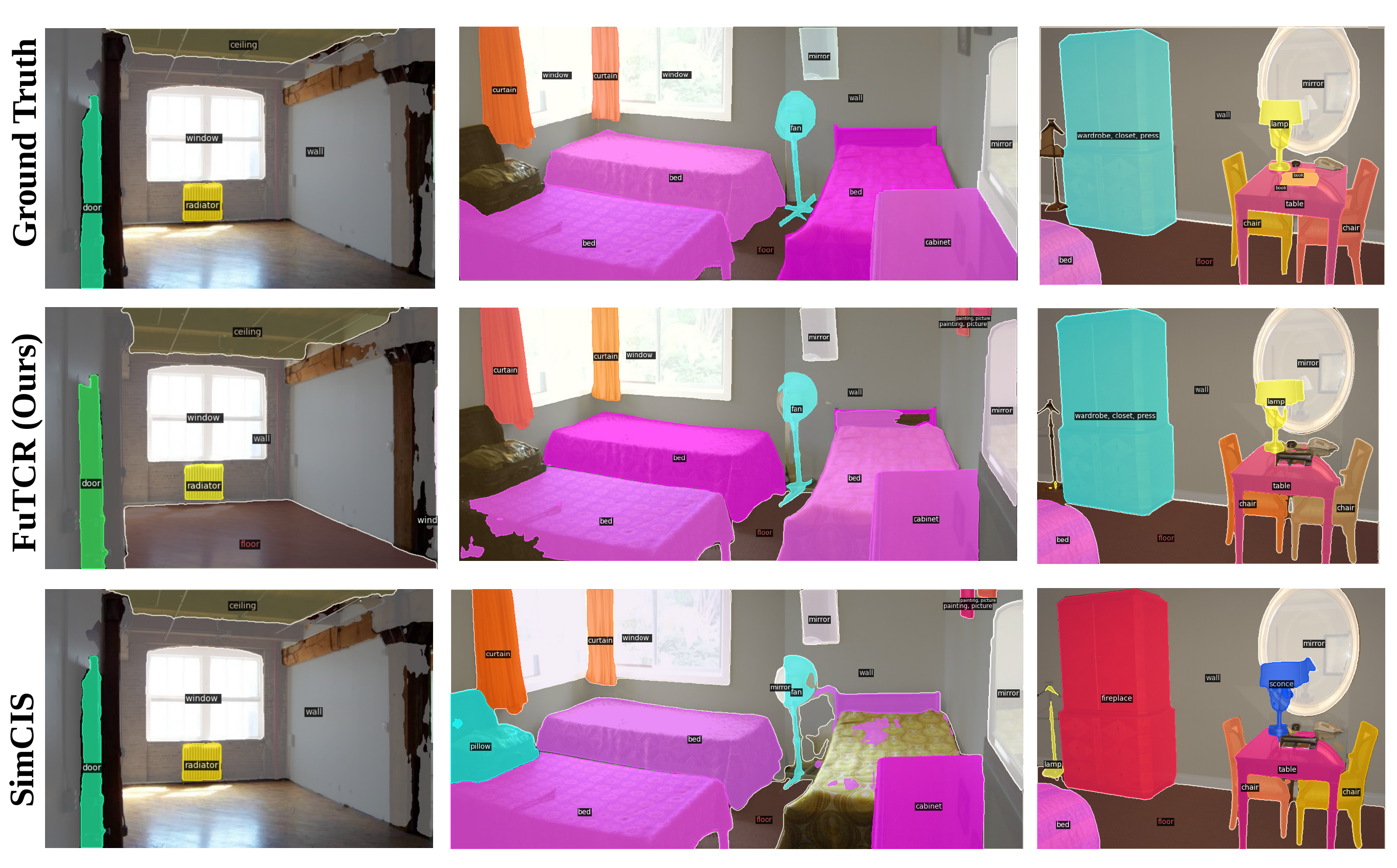}

    \caption{
        Additional qualitative comparisons between FuTCR and SimCIS~\cite{zhu2025rethinking} on ADE20K 100--50.
        The two panels depict diverse scenes where FuTCR recovers more accurate panoptic masks, particularly on newly introduced classes.
    }
    \label{fig:qualitative_appendix_all}
\end{figure*}

In~\cref{fig:qualitative_appendix_all} we provide additional comparisons between our FuTCR and SimCIS~\cite{zhu2025rethinking} on ADE20K 100-50. Across the diverse scenes we selected, we see that SimCIS often either inaccurately predicts labels or produces incomplete masks. Whereas FuTCR yields segmentations that are more closely aligned with the panoptic ground truth. In the third image on the sixth row, for example, SimCIS incorrectly labels the wardrobe as a fireplace. It also fails to properly predict the mask for the 3rd bed in the second image in the sixth row. These patterns illustrate that FuTCR not only improves aggregate metrics as seen in the tables but also produces more reliable and consistent predictions even in challenging layouts.

\section{Dataset and Split Construction}
\label{app:split_construction}
\subsection{Split Construction Details}

We construct ADE20K continual panoptic splits for the 100--5, 100--10, and 100--50 settings. Each setting uses the same base class set of 100 classes and introduces the remaining 50 classes in increments of 5, 10, or 50 classes. The overlap stream corresponds to $(\mathrm{IMG\_OV}, \mathrm{CLS\_OV})=(100,75)$, while the disjoint-image stream corresponds to $(0,75)$.

The base step is shared between the overlap and disjoint-image streams. The overlap stream allows base-step images to reappear during incremental steps, while the disjoint-image stream removes image reuse between base and incremental training. In both streams, the class schedule and evaluation protocol are held fixed.

\begin{table}[h]
\centering
\caption{Number of training images per step for ADE20K 100--5. Step 1 is the shared base step.}
\begin{tabular}{lccccccccccc}
\toprule
Stream & S1 & S2 & S3 & S4 & S5 & S6 & S7 & S8 & S9 & S10 & S11 \\
\midrule
Overlap   & 10177 & 573 & 297 & 617 & 598 & 581 & 941 & 903 & 1475 & 617 & 953 \\
Disjoint  & 10177 & 569 & 297 & 617 & 598 & 570 & 935 & 903 & 1469 & 617 & 953 \\
\bottomrule
\end{tabular}
\end{table}

\subsection{Reduced Incremental Supervision Protocol}

For reduced-data experiments on ADE20K 100--5, we keep the base step, class order, validation set, test set, and number of incremental steps fixed. Only the number of training images per incremental step is reduced. The full setting contains approximately 700 images per incremental step on average. Because the full incremental step sizes vary naturally across class groups, the reduced subsets are sampled proportionally per step rather than forcing each step to contain exactly the same number of images.

\begin{table}[t]
\centering
\caption{
Image counts for reduced incremental supervision on ADE20K 100--5.
The base step is fixed across all reduced-data experiments; reduced supervision is applied only to incremental steps.
}
\resizebox{\textwidth}{!}{
\begin{tabular}{llccccccccccc}
\toprule
Stream & Setting & S1 & S2 & S3 & S4 & S5 & S6 & S7 & S8 & S9 & S10 & S11 \\
\midrule
Overlap 
& $\sim$ 700/step & 10177 & 573 & 297 & 617 & 598 & 581 & 941 & 903 & 1475 & 617 & 953 \\
& $\sim$ 400/step & -- & 327 & 170 & 353 & 342 & 332 & 538 & 516 & 843 & 353 & 545 \\
& $\sim$ 100/step & -- & 80 & 42 & 87 & 84 & 81 & 132 & 127 & 207 & 87 & 134 \\
\midrule
Disjoint
& $\sim$ 700/step & 10177 & 569 & 297 & 617 & 598 & 570 & 935 & 903 & 1469 & 617 & 945 \\
& $\sim$ 400/step & -- & 325 & 170 & 353 & 342 & 326 & 534 & 516 & 839 & 353 & 540 \\
& $\sim$ 100/step & -- & 80 & 42 & 87 & 84 & 80 & 131 & 127 & 206 & 87 & 132 \\
\bottomrule
\end{tabular}
}
\label{tab:reduced_data_counts}
\end{table}

\begin{table}[h]
\centering
\caption{Implementation and FuTCR hyperparameters.}
\begin{tabular}{ll}
\toprule
Parameter & Value \\
\midrule
Backbone & ResNet-50 \\
Segmentation architecture & Mask2Former / SimCIS-style query-based CPS model \\
Framework & Detectron2 \\
Dataset & ADE20K \\
Base classes & 100 \\
Total classes & 150 \\
Increment sizes & 5, 10, 50 \\
Region contrast weight $\lambda_{\mathrm{reg}}$ & 0.5 \\
Repulsion weight $\lambda_{\mathrm{rep}}$ & 0.5 \\
Mask threshold $\tau_{\mathrm{mask}}$ & 0.5 \\
Sampled pixels per region & 70 \\
Temperature $\tau$ & 0.07 \\
Repulsion margin $\gamma$ & 0.0 \\
Optimizer & AdamW \\
Learning rate & $5 \times 10^{-5}$ \\
Batch size & 8 \\
Training iterations 100--5 & 5000 \\
Training iterations 100--10 & 10000 \\
Training iterations 100--50 & 50000 \\
GPUs & 4 NVIDIA RTX A6000 \\
Runtime & $\sim$2 hours per 100--5 incremental step \\
\bottomrule
\end{tabular}
\end{table}

\begin{table}[t]
\centering
\caption{
Ablation study on ADE20K \textbf{100-5} using seed 2. 
We report PQ$_{\mathrm{All}}$ on the validation set under both the overlap setting (100-75) and disjoint setting (0-75). 
The average is computed over the two settings. 
Hyperparameters are selected using validation performance.
}
\label{tab:appendix_ablation}
\resizebox{\textwidth}{!}{
\begin{tabular}{lcccccccccc}
\toprule
Method / Variant 
& $\lambda_{\mathrm{FA}}$ 
& $\tau_{\mathrm{mask}}$ 
& RC 
& PPR.
& KFR 
& $\lambda_{\mathrm{rep}}$ 
& Aux. cls. 
& Val Overl. 
& Val Disj. 
& Val Avg. \\
\midrule
SimCIS~\cite{zhu2025rethinking} 
& -- & -- & -- & -- & -- & -- & -- 
& 31.5 & 30.9 & 31.2 \\

FuTCR w/ aux. cls. 
& 0.5 & 0.5 & \checkmark & 50 & \checkmark & 0.5 & \checkmark 
& 32.5 & 31.6 & 32.1 \\

FuTCR w/o KFR 
& 0.5 & 0.5 & \checkmark & 50 & -- & -- & \checkmark 
& 33.3 & 31.4 & 32.4 \\

FuTCR w/o RC 
& 0.5 & 0.5 & -- & -- & \checkmark & 0.5 & -- 
& 32.9 & 31.6 & 32.3 \\

FuTCR 
& 0.5 & 0.5 & \checkmark & 50 & \checkmark & 0.5 & -- 
& \textbf{33.8} & \textbf{32.0} & \textbf{32.9} \\

\bottomrule
\end{tabular}
}
\end{table}

\section{Auxiliary Prototype Clustering}
\label{subsec:aux_classifier}

Beyond contrast and repulsion, we also experimented with imposing additional structure on future-like region prototypes through an auxiliary clustering branch. We maintain a buffer of region prototypes and periodically run a lightweight $K$-means procedure to estimate cluster centers $\{\mathbf{c}_k\}_{k=1}^{K_{\mathrm{aux}}}$ on the unit sphere. Each prototype $\mathbf{p}_r$ is assigned a pseudo-label $\ell_r = \arg\max_k \mathrm{sim}(\mathbf{p}_r,\mathbf{c}_k)$, and an auxiliary MLP predicts logits $\mathbf{g}_r \in \mathbb{R}^{K_{\mathrm{aux}}}$ from $\mathbf{p}_r$. We then minimize a combination of cross-entropy and a balance term:
\begin{equation}
\mathcal{L}_{\mathrm{aux}}
=
\frac{1}{|\mathcal{R}^{\mathrm{fut}}|}
\sum_{r}
\mathrm{CE}(\mathbf{g}_r, \ell_r)
+
\lambda_{\mathrm{bal}} \,\mathrm{KL}\big(\bar{\mathbf{p}} \,\|\, \mathbf{u}\big),
\label{eq:aux_loss}
\end{equation}
where $\bar{\mathbf{p}}$ is the mean predicted distribution over clusters and $\mathbf{u}$ is the uniform distribution. This head is intended to encourage diverse usage of latent slots for future-like regions and to promote more disentangled prototypes. In our experiments, however, neither this auxiliary classifier nor a related unseen-classifier head yielded consistent gains over our core FuTCR module, so we treat clustering as an \emph{optional} variant and report its effect only in ablations in~\cref{tab:appendix_ablation}.  We find on its own it improved performance, but was harmful when combined with KFR and RC.

\section{Ablation Hyperparameters and Column Definitions}
\label{app:ablation_hyperparams}

\cref{tab:appendix_ablation} provides additional ablations for the FuTCR components and hyperparameters on ADE20K 100--5. 
The columns RC and KFR indicate whether region contrast and known-future repulsion are enabled, respectively. 
RC corresponds to the pixel-to-region contrastive objective that pools confident future-like mask regions into prototypes and encourages sampled pixels from the same region to align with their prototype while separating different region prototypes. 
KFR denotes the known-future repulsion objective, which pushes unlabeled or future-like features away from known-class prototypes to reduce future--old class confusion. 
The mask threshold $\tau_{\mathrm{mask}}$ controls which pixels are included when forming a region prototype, while ``PPR'' (Pixels per region) specifies the number of sampled pixel features used from each selected region. 
The coefficient $\lambda_{\mathrm{FA}}$ weights the future-aware branch, and $\lambda_{\mathrm{rep}}$ controls the strength of the repulsion term. 
The ``Aux. cls.'' column indicates whether the optional auxiliary clustering classifier is enabled; this branch was explored as an additional way to organize future-like prototypes, but was not used in the final model because it did not consistently improve validation performance.

We report validation PQ$_{\mathrm{All}}$ under both controlled data streams. 
The 100--75 stream denotes the overlap setting, where base-step images may reappear in later incremental steps, while 0--75 denotes the disjoint-image setting, where base-step and incremental-step images are disjoint but the class schedule is kept fixed. 
Following our experimental protocol, hyperparameters are selected using validation performance only; test results are shown for transparency and are not used for model selection.

The ablation results show that the selected FuTCR configuration is chosen by validation performance. Removing the auxiliary classifier improves the validation average compared with the full auxiliary variant, suggesting that the main gains come from the combination of region contrast and known-future repulsion rather than from auxiliary prototype clustering. KFR is especially important because it directly targets the collapse of future-like regions into known-class decision regions, while RC is most effective when paired with KFR.

\section{Training Overview.}
As detailed in~\cref{alg:futcr_appendix}, at each incremental step $t$, FuTCR first identifies query regions that predominantly cover unlabeled pixels and treats them as potential future-class regions. These regions are used to form prototypes that supervise a future-targeted contrastive objective in the shared query feature space, while an additional repulsion term pushes unlabeled features away from known-class prototypes. The final objective combines standard panoptic supervision with these future-aware contrastive and repulsive regularizers.

\begin{algorithm}[t]
\caption{FuTCR Training at Incremental Step $t$}
\label{alg:futcr_appendix}
\begin{algorithmic}[1]
\State \textbf{Input:} Query-based panoptic model $f_{\theta_t}$ (backbone $f_{\text{backbone}}$, decoder $f_{\text{dec}}$, classifier weights $W_t$);
incremental dataset $\mathcal{D}_t$ with current classes $\mathcal{C}^t$ and known classes $\mathcal{C}^{\le t}$;
mask threshold $\tau_{\mathrm{mask}}$, contrast temperature $\tau$, repulsion margin $\gamma$;
loss weights $\lambda_{\mathrm{reg}}, \lambda_{\mathrm{rep}}$.
\State \textbf{Output:} Updated parameters $\theta_t$.

\For{each minibatch $\{(x_b,y_b)\}_{b=1}^{B} \subset \mathcal{D}_t$}
  \State \textit{/* Supervised panoptic learning on $\mathcal{C}^t$ */}
  \For{$b = 1$ to $B$}
    \State $F_b \gets f_{\text{backbone}}(x_b)$
    \State $\mathbf{h}_t(x_b) \gets f_{\text{dec}}(F_b)$
    \State $\{\hat{m}_{b,q}\}_{q=1}^{Q} \gets f_{\theta_t}(x_b)$
  \EndFor
  \State Compute panoptic loss $\mathcal{L}_{\mathrm{pan}}$ on $\mathcal{C}^t$ using $(\hat{m}_{b,q}, y_b)$

  \State \textit{/* Future-region discovery */}
  \State $\mathcal{R}^{\mathrm{fut}} \gets \emptyset$
  \For{$b = 1$ to $B$}
    \State Identify unlabeled pixels in $y_b$ w.r.t.\ $\mathcal{C}^{\le t}$
    \For{$q = 1$ to $Q$}
      \If{$\hat{m}_{b,q}$ is confident, sufficiently large, and majority-supported on unlabeled pixels}
        \State $\mathcal{R}^{\mathrm{fut}} \gets \mathcal{R}^{\mathrm{fut}} \cup \{(b,q)\}$
      \EndIf
    \EndFor
  \EndFor

  \State \textit{/* Future-targeted region contrast */}
  \State $\mathcal{P} \gets \emptyset$ \Comment{region prototypes}
  \State Collect pixel anchors $\{\mathbf{f}_n\}_{n=1}^{N}$
  \For{each $r = (b,q) \in \mathcal{R}^{\mathrm{fut}}$}
    \State $\Omega_r \gets \{(i,j) \mid \hat{m}_{b,q}(i,j) > \tau_{\mathrm{mask}}\}$
    \State $\mathbf{p}_r \gets \frac{1}{|\Omega_r|} \sum_{(i,j)\in\Omega_r} F_b(:, i, j)$
    \State Sample pixel features from $\Omega_r$ as anchors $\{\mathbf{f}_n\}_{n:r(n)=r}$
    \State $\mathcal{P} \gets \mathcal{P} \cup \{\mathbf{p}_r\}$
  \EndFor
  \State Compute region-level InfoNCE loss $\mathcal{L}_{\mathrm{reg}}$ over $\{\mathbf{f}_n\}$ and $\mathcal{P}$ with temperature $\tau$

  \State \textit{/* Known-class repulsion on unlabeled pixels */}
  \State Collect unlabeled pixel features $\{\mathbf{z}_u\}_{u\in\mathcal{I}^{\mathrm{unlb}}}$ from $\{F_b\}$
  \State Let $\{\mathbf{w}_c\}_{c\in\mathcal{C}^{\le t}}$ be normalized prototypes derived from $W_t$
  \For{each $u \in \mathcal{I}^{\mathrm{unlb}}$}
    \State $c^{\star}(u) \gets \arg\max_{c\in\mathcal{C}^{\le t}} \mathrm{sim}(\mathbf{z}_u,\mathbf{w}_c)$
  \EndFor
  \State Compute repulsion loss $\mathcal{L}_{\mathrm{rep}}$ with margin $\gamma$ using $\{\mathbf{z}_u\}$ and $\{c^{\star}(u)\}$

  \State \textit{/* Final objective and update */}
  \State $\mathcal{L}_{\mathrm{total}} \gets \mathcal{L}_{\mathrm{pan}}
            + \lambda_{\mathrm{reg}}\,\mathcal{L}_{\mathrm{reg}}
            + \lambda_{\mathrm{rep}}\,\mathcal{L}_{\mathrm{rep}}$
  \State Update $\theta_t$ via backpropagation on $\mathcal{L}_{\mathrm{total}}$
\EndFor
\end{algorithmic}
\end{algorithm}

\section{Impact Statement.}
\label{app:impact_statement}
FuTCR advances continual representation learning for long-lived perception systems, with potential benefits for robotics, assistive sensing, and autonomous systems where models must adapt to newly observed categories over time. However, improved continual perception could also be used in surveillance or other monitoring applications, and failures in safety-critical settings could lead to incorrect recognition of newly introduced objects. We therefore view FuTCR as a research contribution rather than a deployment-ready system, and recommend careful evaluation under domain shift, fairness, and safety constraints before real-world use.

\section{Datasets, Codebases, and Licenses.}
\label{app:data_code_license}
We conduct experiments on ADE20K~\cite{zhou2017sceneade20k}, following the continual panoptic segmentation protocols of prior work. ADE20K is publicly available for research use under its dataset terms. Our implementation builds on Mask2Former~\cite{cheng2022masked} and Detectron2, which are publicly released research codebases. We also compare against and/or adapt publicly available implementations of continual panoptic segmentation baselines including SimCIS~\cite{zhu2025rethinking}, CoMBO~\cite{fang2025combo}, BalConpas~\cite{chen2024strike}, and Eclipse~\cite{kim2024eclipse}. We cite the original papers and use these assets according to their released licenses and terms of use.

\end{document}